\documentclass[runningheads]{llncs}

\usepackage[mobile]{eccv}

\usepackage{eccvabbrv}

\usepackage{graphicx}
\usepackage{booktabs}

\usepackage[breaklinks,colorlinks,citecolor=eccvblue]{hyperref}

\usepackage{algorithm} 
\usepackage{algorithmic}  
\usepackage[algo2e,ruled,linesnumbered]{algorithm2e}
\SetKwInOut{Parameter}{parameter}
\usepackage{booktabs}
\usepackage{multirow}
\usepackage{svg}
\usepackage[title]{appendix}

\newcommand*\samethanks[1][\value{footnote}]{\footnotemark[#1]}

\begin{document}

\title{ScaleDreamer: Scalable Text-to-3D Synthesis with Asynchronous Score Distillation} 

\titlerunning{ScaleDreamer}

\author{Zhiyuan Ma\inst{1,2} \and
Yuxiang Wei\inst{1,5} \and
Yabin Zhang\inst{1} \and \\
Xiangyu Zhu\inst{3,4} \and
Zhen Lei\inst{1,2,3,4\thanks{Corresponding authors. }} \and
Lei Zhang\inst{1\samethanks}
}

\authorrunning{Ma et al.}

\institute{The Hong Kong Polytechnic University, PolyU \and 
Center for Artificial Intelligence and Robotics, HKISI CAS \and
State Key Laboratory of Multimodal Artificial Intelligence Systems, CASIA \and
School of Artificial Intelligence, University of Chinese Academy of
Sciences, UCAS \and
Harbin Institute of Technology, HIT\\
\url{https://github.com/theEricMa/ScaleDreamer}}

\maketitle




\begin{abstract}
    By leveraging the text-to-image diffusion priors, score distillation can synthesize 3D contents without paired text-3D training data. Instead of spending hours of online optimization per text prompt, recent studies have been focused on learning a text-to-3D generative network for amortizing multiple text-3D relations, which can synthesize 3D contents in seconds. However, existing score distillation methods are hard to scale up to a large amount of text prompts due to the difficulties in aligning pretrained diffusion prior with the distribution of rendered images from various text prompts. Current state-of-the-arts such as Variational Score Distillation finetune the pretrained diffusion model to minimize the noise prediction error so as to align the distributions, which are however unstable to train and will impair the model's comprehension capability to numerous text prompts. Based on the observation that the diffusion models tend to have lower noise prediction errors at earlier timesteps, we propose Asynchronous Score Distillation (ASD), which minimizes the noise prediction error by shifting the diffusion timestep to earlier ones. ASD is stable to train and can scale up to 100k prompts. It reduces the noise prediction error without changing the weights of pre-trained diffusion model, thus keeping its strong comprehension capability to prompts. We conduct extensive experiments across different 2D diffusion models, including Stable Diffusion and MVDream, and text-to-3D generators, including Hyper-iNGP, 3DConv-Net and Triplane-Transformer. The results demonstrate ASD's effectiveness in stable 3D generator training, high-quality 3D content synthesis, and its superior prompt-consistency, especially under large prompt corpus.

\end{abstract}

\begin{figure*}[!t]
    \centering
    \includegraphics[width=1\textwidth]{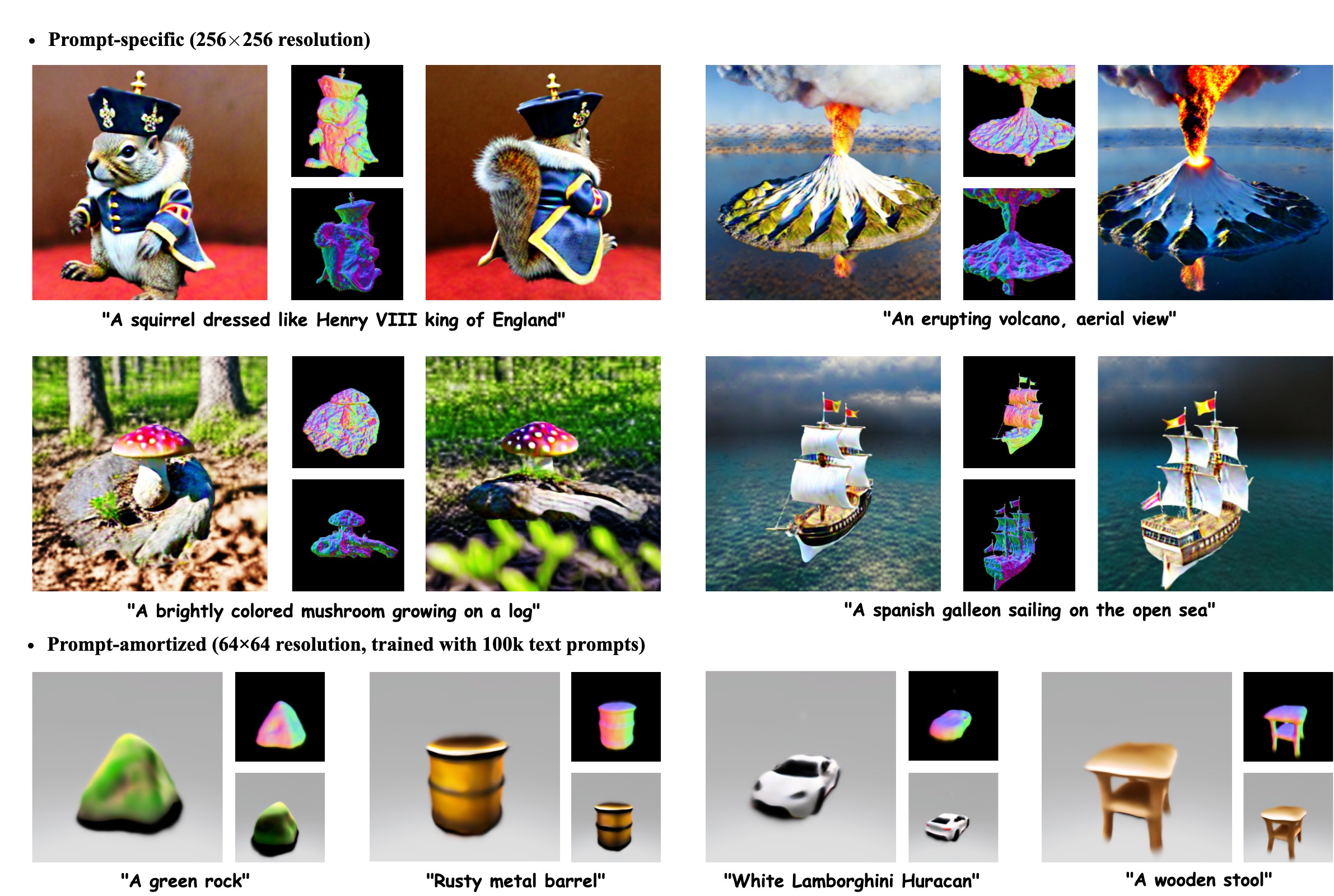}
    \caption{\textbf{Top two rows}: Asynchronous Score Distillation (ASD) for prompt-specific text-to-3D generation. \textbf{Bottom row}: ASD for prompt-amortized generation, which learns a text-to-3D generator on multiple prompts without 3D ground truths. ASD has strong capability to scale up the training corpus to as much as \textbf{100k} text prompts.}
    \label{fig:teaser}
\end{figure*}

\section{Introduction}
\label{sec:intro}


Text-to-3D aims to generate realistic 3D contents from the given textual descriptions~\cite{poole2022dreamfusion}, which is particularly useful in many applications such as virtual reality~\cite{wohlgenannt2020virtual} and game design~\cite{koster2013theory}. The main challenge of this task, however, lies in how to generate high-quality 3D contents conditioned on the abstract and diverse textual descriptions.
Many existing text-to-3D methods \cite{poole2022dreamfusion,wang2023score,wang2023prolificdreamer,lin2023magic3d,metzer2023latent,guo2023stabledreamer,ma2023x,zhou2023dreampropeller,qiu2023richdreamer,li2023mvcontrol,liang2023luciddreamer,li2023sweetdreamer,liu2023unidream,ding2023text} are optimization-based ones, which distill the guidance from the powerful pretrained text-to-image diffusion models~\cite{rombach2022high,balaji2022ediffi,li2023sweetdreamer,qiu2023richdreamer,liu2023unidream,ding2023text,zhao2023efficientdreamer} via score distillation~\cite{poole2022dreamfusion,wang2023prolificdreamer,yu2023text,wu2024one}. In general, these methods employ the KL divergence to reduce the discrepancy between the distribution of rendered images and the desired image distribution embedded in the 2D diffusion prior, while they differ in how to use the pretrained diffusion prior to model the distribution of rendered images. Extensive efforts have been made to explore prompt-specific optimization of various 3D representations, including implicit radiance fields~\cite{poole2022dreamfusion}, explicit radiance fields~\cite{metzer2023latent, lin2023magic3d,wang2023prolificdreamer}, DmTets~\cite{tsalicoglou2023textmesh,zhao2024flexidreamer} and 3D Gaussians~\cite{chen2023text}. 
Typically, tens of minutes to hours are needed to optimize a single 3D representation for one prompt to achieve the desired result.


Compared to the aforementioned optimization-based text-to-3D methods, learning-based methods~\cite{liu2023pi3d,jun2023shap,cao2023large,tang2023volumediffusion,ren2023xcube,mercier2024hexagen3d,xie2024latte3d} can largely reduce the computational cost by training a text-conditioned 3D generative network. 
With the availability of 3D object collections~\cite{wu2023omniobject3d,deitke2023objaverse,yu2023mvimgnet}, a deep network can be trained in a supervised manner so that 3D outputs can be generated in several seconds. Unfortunately, the size of existing text-3D datasets is far from sufficient compared to text-image datasets~\cite{schuhmann2022laion}, limiting the text-to-3D generation performance of trained models. Inspired by the optimization-based text-to-3D methods that use pretrained 2D diffusion models, efforts have been made to train text-to-3D networks by using 2D diffusion models as supervisors~\cite{lorraine2023att3d,qian2024atom,xie2024latte3d} without using text-3D pairs. For example, a text-conditioned 3D hyper-network is trained in ATT3D~\cite{lorraine2023att3d} via Score Distillation Sampling (SDS) \cite{poole2022dreamfusion}. Nevertheless, this method suffers from numerical instability, which has been observed in subsequent studies\cite{qian2024atom,xie2024latte3d} that apply SDS to different 3D generator networks.


Despite the success of score distillation in optimization-based text-to-3D generation~\cite{poole2022dreamfusion,yu2023text,wang2023prolificdreamer}, its application to learning-based text-to-3D frameworks is rather limited because of the unstable training or unsatisfactory results.
We argue that the primary challenge lies in how to efficiently and effectively leverage the pretrained 2D diffusion prior to represent the distribution of images rendered by the 3D generator. For example, SDS~\cite{poole2022dreamfusion} forces the rendered images to adhere to the Dirac distribution, which causes numerical instability in 3D generator training \cite{lorraine2023att3d,xie2024latte3d}. Variational Score Distillation (VSD)~\cite{wang2023prolificdreamer} finetunes the 2D diffusion prior for distribution alignment via minimizing the noise prediction error. However, the finetuning changes the pretrained diffusion network and hurts its comprehension capability to numerous text prompts, leading to mode collapse when the size of text prompts is extended. 

To address the above mentioned issues, we propose Asynchronous Score Distillation (ASD). Like VSD, ASD aims to minimize the noise prediction error. Different from VSD, ASD does not finetune the pretrained 2D diffusion network; instead, it achieves the goal by shifting the diffusion timestep. This is based on the observation that diffusion networks will have smaller noise prediction errors in earlier timesteps~\cite{yang2023eliminating}; therefore, we can shift the timestep to an earlier step to achieve a similar goal to VSD, \ie, reducing the noise prediction error. In this way, the diffusion network can be frozen in training and its strong text comprehension capability can be well-preserved. 
The shifted timesteps can be well sampled from a pre-defined range for most prompts. To evaluate the performance of ASD, we conduct extensive experiments by using three types of generator architectures,  \ie Hyper-iNGP~\cite{lorraine2023att3d}, 3DConv-Net~\cite{bahmani2023cc3d} and Triplane-Transformer ~\cite{hong2023lrm}, and two types of 2D diffusion models, \ie, Stable Diffusion~\cite{rombach2022high} and MVDream~\cite{shi2023mvdream}, across various prompt corpus sizes.  We conduct extensive experiments to evaluate the superiority of ASD to previous methods, including the stable training of 3D generators, the production of high-quality 3D outputs, the high content fidelity to input prompts, as well as its scalability to larger corpus sizes, \eg, 100k prompts. Some results are shown in Fig.~\ref{fig:teaser}.

\section{Literature Review}
\subsection{Text-to-3D with Score Distillation}
Text-to-3D takes text description, a.k.a. text prompt $y$, as input, and outputs 3D representation $\theta$ that renders high-fidelity images at any camera view $\pi$. 
Thanks to the powerful text-to-image diffusion models~\cite{rombach2022high,zhao2023efficientdreamer,shi2023mvdream,liu2023unidream,qiu2023richdreamer}, we can optimize $\theta$ to align with $y$ by computing the objective $\mathcal{L}(\boldsymbol{x}, y)$ on the rendered image $\boldsymbol{x} = g(\theta, \pi)$ from camera view $\pi$. Through differential rendering, $\theta$ can be updated with the gradient $\nabla_\theta \mathcal{L}(\theta, y) = \frac{\partial \mathcal{L}(\boldsymbol{x}, y)}{\partial \boldsymbol{x}} \frac{\partial \boldsymbol{x}}{\partial \theta}$. This technique is generally termed as score distillation. Unlike data-driven techniques~\cite{liu2023pi3d,jun2023shap,cao2023large,tang2023volumediffusion,ren2023xcube}, score distillation approaches \cite{poole2022dreamfusion,yu2023text,wang2023prolificdreamer,tang2023dreamgaussian,chen2023fantasia3d,lin2023magic3d,huang2023dreamtime,lee2024dreamflow} can produce high-quality 3D content without the need for 3D training datasets.

 \textbf{Prompt-Specific Text-to-3D}. Existing score distillation methods~\cite{poole2022dreamfusion,yu2023text,wang2023prolificdreamer} were originally developed to output a single 3D result $\theta$ for a single text prompt $y$ via online optimization: $\min_{\theta}\mathbb{E}_{\pi, \boldsymbol{x}=g(\theta, \pi)}{\left[\mathcal{L}(\boldsymbol{x}, y)\right]}$. The utilized 3D representations, \eg, NeRF~\cite{poole2022dreamfusion,muller2022instant}, DmTet~\cite{shen2021deep,zhao2024flexidreamer}, and 3D Gaussian~\cite{tang2023dreamgaussian,yi2023gaussiandreamer,vilesov2023cg3d,jiang2024brightdreamer,lin2024dreampolisher,tang2023stable}, are not designed to render scenes from varying text prompts. Therefore, the optimization has to be conducted again for newly provided text prompts. The optimization process typically costs tens of minutes to hours. 

\textbf{Prompt-Amortized Text-to-3D}. To mitigate the computational costs in prompt-specific methods, recent studies \cite{lorraine2023att3d,li2023instant3d,qian2024atom,xie2024latte3d} have attempted to use score distillation to train a text-to-3D generator $\theta = \mathcal{G}(y)$, aiming to generate multiple 3D representations from a set of text prompts $S_y = \{ y \}$. These methods can generate 3D results from queried text prompt in seconds. As proposed by ATT3D~\cite{lorraine2023att3d}, the 3D generator training is performed by minimizing   $\min_{\mathcal{G}}\mathbb{E}_{\pi, y \in S_y, \boldsymbol{x}=g(\mathcal{G}(y), \pi)}{\left[\mathcal{L}(\boldsymbol{x}, y)\right]}$ over all text prompts.  Unlike data-driven approaches \cite{hong2023lrm,tang2024lgm,xu2023dmv3d}, score distillation bypasses the scarcity of text-3D data pairs because the 2D diffusion prior can offer the guidance to align the 3D output with the input text prompt. However, its application is currently restricted to training the 3D generator within a limited range of text prompts.

\subsection{Representative Score Distillation Methods}
\label{sec:review_score_distillation}

Denote by $\phi$ the 2D diffusion prior~\cite{rombach2022high,shi2023mvdream} and by $p^{\phi}\left(\boldsymbol{x} \mid y\right)$ the text-conditioned image distribution embedded within $\phi$, the objectives of most existing score distillation methods can be generally concluded as minimizing the objective
$
    \mathcal{L}(\theta, y) = \mathbb{E}_{\pi,t,\boldsymbol{\epsilon}, \boldsymbol{x} = g\left(\theta, \pi\right)}\left[\omega(t) D_{\mathrm{KL}}\left(
        q_t^{\theta}\left(\boldsymbol{x}_{t} \mid \pi \right) \|
        p^\phi_t\left(\boldsymbol{x}_{t} \mid y^\pi\right)
    \right) \right],
$
where $D_{\mathrm{KL}}$ denotes KL divergence, $q_t^\theta\left(\boldsymbol{x}_{t} \mid \pi \right)$ denotes the distribution of images $\boldsymbol{x}$ rendered at camera view $\pi$ at diffusion timestep $t$~\cite{ho2020denoising}, and the same for $p^\phi_{t}(\boldsymbol{x}_{t} \mid y)$. $\omega(t)$ is a timestep-dependent weight~\cite{poole2022dreamfusion}. $y^\pi$ denotes the view-dependent strategy~\cite{rombach2022high} or view-awareness ~\cite{shi2023mvdream,qiu2023richdreamer} to prompt the different camera views~\cite{poole2022dreamfusion}. To minimize this objective, the gradient w.r.t. $\theta$ can be calculated as per~\cite{wang2023prolificdreamer}:
\begin{equation}
   \nabla_\theta\mathcal{L}(\theta, y) \!=\! \mathbb{E}_{\pi\!,t\!, \boldsymbol{\epsilon}\!}\!\left[ \!\omega(t)\!\left(  
    \!\underbrace{\!
        -\sigma_t \nabla_{\boldsymbol{x}_t} \log p^\phi_t\!\left(\boldsymbol{x}_t \!\mid\! y^\pi\right)
        \!}_{ \boldsymbol{\epsilon}_\phi\left( \boldsymbol{x}_t ; t, y^{\pi} \right)}
    -\underbrace{\!
        \!\left(\!-\sigma_t \nabla_{\boldsymbol{x}_t} \log q_t^{\theta}\!\left(\boldsymbol{x}_t \!\mid\! \pi \right)\!\right)
        \!}_{ \boldsymbol{\epsilon}_{\theta}\left(\boldsymbol{x}_t;t,\pi,y\right)}
    \!\right)\!\frac{\partial \boldsymbol{x}}{\partial \theta} \!\right]\!,
\label{equ:contrastive_score_models}
\end{equation}
where the first term $-\sigma_t \nabla_{\boldsymbol{x}_t} \log p_t^\phi\left(\boldsymbol{x}_t \mid y^\pi\right)$ corresponds to the score function~\cite{song2020score} of the desired image distribution, and it can be achieved by predicting the noise $\boldsymbol{\epsilon} \sim \mathcal{N}\left(0, \boldsymbol{I}\right)$ in the noisy image $\boldsymbol{x}_t = \alpha_t \boldsymbol{x} + \sigma_t \boldsymbol{\epsilon}$ using the pretrained 2D diffusion model $\boldsymbol{\epsilon}_\phi\left(\boldsymbol{x}_t ; t, y^\pi\right)$~\cite{rombach2022high,shi2023mvdream}. Existing score distillation methods \cite{poole2022dreamfusion,wang2023prolificdreamer,yu2023text} mainly differ in how to model 
 $-\sigma_t \nabla{\boldsymbol{x}_t} \log q_t^{\theta}\left(\boldsymbol{x}_t \mid \pi \right)$, which corresponds to the score function of the distribution of rendered images $q^\theta\left(\boldsymbol{x} \mid \pi \right)$. We denote this term in Eq.~\ref{equ:contrastive_score_models} as $\boldsymbol{\epsilon}_{\theta}\left(\boldsymbol{x}_t;t,\pi,y\right)$ in the following context, since it represents a diffusion model that corresponds to $\theta$. A summary of the objectives of major score distillation methods is shown in  Table~\ref{table:formulas}.

 The objective of \textbf{Score Distillation Sampling} (SDS)~\cite{poole2022dreamfusion} is 
$
    \nabla_\theta\mathcal{L}_{\mathrm{SDS}}(\theta, y) \triangleq \mathbb{E}_{\pi, t, \boldsymbol{\epsilon}}\left[
        \omega(t)\left(
            \boldsymbol{\epsilon}_\phi\left(\boldsymbol{x}_t ; t, y^\pi\right) -
            \boldsymbol{\epsilon}
        \right)  \frac{\partial \boldsymbol{x}}{\partial \theta}
    \right],
$
which approximates the term $\boldsymbol{\epsilon}_{\theta}\left(\boldsymbol{x}_t;t,\pi,y\right)$ in Eq.~\ref{equ:contrastive_score_models} as the ground-truth noise $\boldsymbol{\epsilon}$. That is, SDS assumes that $q^{\theta}\left(\boldsymbol{x} \mid \pi \right)$ adheres to a Dirac distribution $\delta\left(\boldsymbol{x} - g\left(\theta, \pi\right)\right)$~\cite{wang2023prolificdreamer}, which is characterized by a non-zero density at the singular point of $\boldsymbol{x} = g(\theta,\pi)$ and zero density everywhere else. However, updating $\theta$ under the Dirac distribution might be troublesome~\cite{wang2023prolificdreamer}. We may need to set the CFG (Classifier Free Guidance)~\cite{ho2022classifier} as high as 100 for model convergence, which will produce excessively large gradients and lead to unstable optimization. This problem is alleviated by \textbf{Classifier Score Distillation} (CSD)~\cite{yu2023text}, which uses the classifier component~\cite{ho2022classifier} in SDS as the objective:  
$
    \nabla_\theta\mathcal{L}_{\mathrm{CSD}}(\theta, y) \triangleq \mathbb{E}_{\pi, t, \boldsymbol{\epsilon}}\left[
        \omega(t)\left(
            \boldsymbol{\epsilon}_\phi\left(\boldsymbol{x}_t ; t, y^\pi\right) -
            \boldsymbol{\epsilon}_\phi\left(\boldsymbol{x}_t ; t\right)
        \right)  \frac{\partial \boldsymbol{x}}{\partial \theta}
    \right].
$
CSD can be regraded as straightforwardly using the unconditional term of the diffusion prior $\boldsymbol{\epsilon}_\phi\left(\boldsymbol{x}_t ; t\right)$ to represent $\boldsymbol{\epsilon}_{\theta}\left(\boldsymbol{x}_t;t,\pi,y\right)$ in Eq.~\ref{equ:contrastive_score_models}. Unfortunately, in the case of prompt-amortized training, this term may not provide effective gradient, because $\boldsymbol{\epsilon}_\phi\left(\boldsymbol{x}_t ; t\right)$ is unconditional to the provided text-prompts. 
In contrast, \textbf{Variational Score Distillation} (VSD)~\cite{wang2023prolificdreamer} models $\boldsymbol{\epsilon}_{\theta}\left(\boldsymbol{x}_t;t,\pi,y\right)$ with  another text-aware diffusion model $\boldsymbol{\epsilon}_{\phi^{\prime}}\left(\mathbf{x}_t ; t, \pi, y\right) $, leading to
$
    \nabla_\theta\mathcal{L}_{\mathrm{VSD}}(\theta, y) \triangleq \mathbb{E}_{\pi, t, \boldsymbol{\epsilon}}\left[
        \omega(t)\left(
            \boldsymbol{\epsilon}_\phi\left(\boldsymbol{x}_t ; t, y^\pi\right) -
            \boldsymbol{\epsilon}_{\phi^{\prime}}\left(\mathbf{x}_t ; t, \pi, y\right)
        \right)  \frac{\partial \boldsymbol{x}}{\partial \theta}
    \right],
$
where $\boldsymbol{\epsilon}_{\phi^{\prime}}\left(\mathbf{x}_t ; t, \pi, y\right)$ is achieved by finetuning the pretrained 2D diffusion prior $\boldsymbol{\epsilon}_\phi\left(\boldsymbol{x}_t ; t, y^\pi\right)$ to align with the rendered image distribution $q^\theta(\boldsymbol{x} \mid \pi)$ via parameter efficient adaptation~\cite{hu2021lora}. In practice, this is conducted by alternatively optimizing $\theta$ and finetuning $\phi$ with the noise prediction objective $\|\boldsymbol{\epsilon}_{\phi}\left(\boldsymbol{x}_t ; t, y\right) - \boldsymbol{\epsilon}\|^2_2$~\cite{ho2020denoising} such that:
\begin{equation}
    \mathbb{E}_{\pi, t, \boldsymbol{\epsilon}} \left[ 
        \|
            \boldsymbol{\epsilon}_{\phi\prime}\left(\boldsymbol{x}_t ; t, \pi, y\right) - \boldsymbol{\epsilon}
        \|^2_2 
    \right] \leq 
    \mathbb{E}_{\pi, t, \boldsymbol{\epsilon}} \left[ 
        \|
            \boldsymbol{\epsilon}_\phi\left(\boldsymbol{x}_t ; t, y^\pi\right) - \boldsymbol{\epsilon}
        \|^2_2
    \right].
    \label{equ:vsd_finetuneing_objective}
\end{equation}
The above equation reveals that \textbf{a better alignment with the distribution of $q^\theta(\boldsymbol{x} \mid \pi)$ can be achieved by a more accurate noise prediction}.

While VSD achieves state-of-the-art results in prompt-specific text-to-3D~\cite{wang2023prolificdreamer,he2023t}, it changes the diffusion prior's parameters by alternately optimizing $\theta$ and finetuning $\phi$. This forms a bi-level optimization, known to be problematic in generative adversarial training~\cite{thanh2020catastrophic}, and may be troublesome for training prompt-amortized text-to-3D models, because the change of pre-trained diffusion model might impairs its comprehension capability on a wide range of text-prompts. In specific, the pre-trained 2D diffusion model may have to sacrifice its generation capability in order to align with the distribution of rendered images, making it fail to produce good gradient for training the 3D generator.

\begin{table}[!t]
\setlength{\tabcolsep}{0.18cm}
\centering
\begin{tabular}{|c|c|}
\hline
\textbf{Method} & \textbf{Gradient of $\mathcal{L}(\boldsymbol{x}, y)$ \wrt $\boldsymbol{x} = g\left(\theta, \pi\right)$} \\ \hline
SDS~\cite{poole2022dreamfusion}  & $\mathbb{E}_{t, \boldsymbol{\epsilon}}\left[\omega(t) \left(\boldsymbol{\epsilon}_{\phi}\left(\boldsymbol{x}_t; t, y^{\pi}\right) - {\color{Blue}\boldsymbol{\epsilon}}\right)\right]$ \\ \hline
CSD~\cite{yu2023text}     & $\mathbb{E}_{t, \boldsymbol{\epsilon}}\left[\omega(t)\left(\boldsymbol{\epsilon}_{\phi}\left(\boldsymbol{x}_t; t, y^{\pi}\right) - {\color{Blue}\boldsymbol{\epsilon}_{\phi}\left(\boldsymbol{x}_{t}, t\right)}\right)\right]$ \\ \hline
VSD~\cite{wang2023prolificdreamer}     & $\mathbb{E}_{t, \boldsymbol{\epsilon}}\left[\omega(t)\left(\boldsymbol{\epsilon}_{\phi}\left(\boldsymbol{x}_t; t, y^{\pi}\right) - {\color{Blue}\boldsymbol{\epsilon}_{\phi'}\left(\boldsymbol{x}_t; t, \pi, y\right)}\right)\right] $          \\ \hline
ASD  
(Ours)     & $\mathbb{E}_{t, \boldsymbol{\epsilon} }\left[\omega(t)\left(\boldsymbol{\epsilon}_{\phi}\left(\boldsymbol{x}_t; t, y^{\pi}\right) - {\color{Blue}\boldsymbol{\epsilon}_{\phi}\left(\boldsymbol{x}_{t+\Delta t}; t+\Delta t, y^{\pi}\right)}\right)\right] $ \\ \hline
\end{tabular}
\vspace{+2mm}
\caption{Objectives of representative score distillation methods. ASD introduces  $\Delta t$ alongside $t$ to align with the rendered image distribution  $q^\theta(\boldsymbol{x} \mid \pi)$. }
\label{table:formulas}
\end{table}

\vspace{-2.mm}
\section{Asynchronous Score Distillation (ASD)}
\vspace{-1.mm}
\subsection{Objective of ASD}
\vspace{-1.mm}
\label{se:objective_of_asd}

From the above discussions in Sec. \ref{sec:review_score_distillation}, it can be seen that one key issue in VSD is to minimize the noise prediction error so that the model output can be aligned with the desired distribution of rendered images. VSD achieves this goal via finetuning the pre-trained 2D diffusion model, which however sacrifices its comprehension capability on text prompts. One interesting question is: can we minimize the noise prediction error without changing the pre-trained diffusion network weights? Fortunately, we find that this is possible and in this section we present a new objective function to achieve this goal. 

Recall that diffusion models solve the stochastic differential equation~\cite{song2020score} via reversing the noise added along different stages, a.k.a. diffusion timestep $t \in \{T_{\mathrm{max}}, \dots, T_{\mathrm{min}}\}$ via $\boldsymbol{x}_t = \alpha_t \boldsymbol{x} + \sigma_t \boldsymbol{\epsilon} $~\cite{ho2020denoising}. The influence of the noise $\boldsymbol{\epsilon} \sim \mathcal{N}(0,  \boldsymbol{I})$ on the image $\boldsymbol{x}$ is incrementally reduced as the process progresses from the initial timestep $T_\mathrm{max}$ to the final timestep $T_{\mathrm{min}}$, which is controlled by the scalars $\alpha_t$ and $\sigma_t$. Consequently, the diffusion model's noise prediction accuracy will vary with the timestep $t$, at which the identical noise $\boldsymbol{\epsilon}$ is added. 
To evaluate this, we consider a diffusion model with fixed image $\boldsymbol{x}$, noise $\boldsymbol{\epsilon}$ and condition $y$, but varied timestep $t$. We denote such a diffusion model as $\boldsymbol{\epsilon}(t)$ and explore how its prediction error, denoted by  $e(t)$=$\|\boldsymbol{\epsilon}(t) - \boldsymbol{\epsilon}\|^2_2$, changes with $t$. 

\begin{figure*}[!t]
    \centering
    \includegraphics[width=0.8\textwidth]{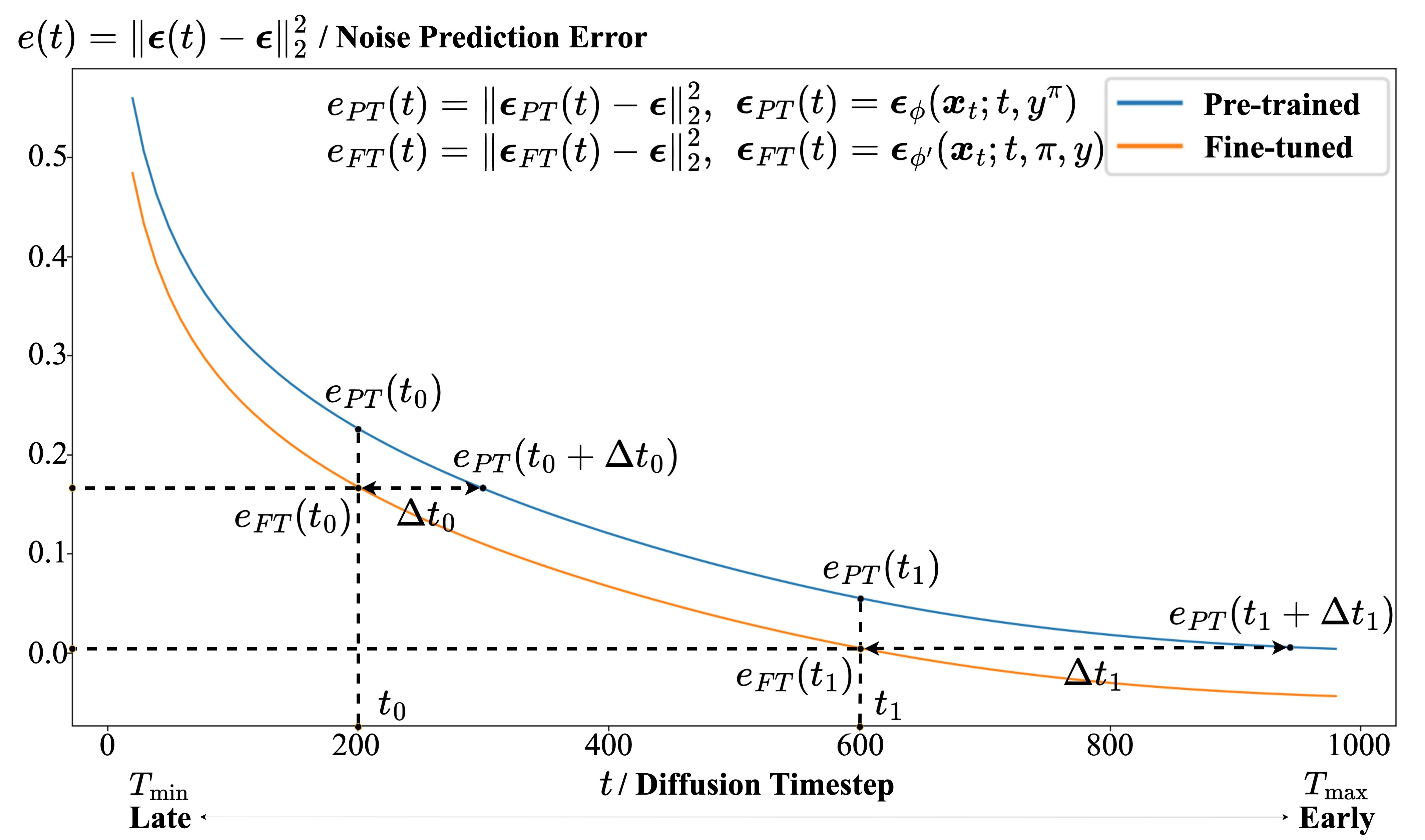}
    \caption{Illustration of the noise prediction error of the pre-trained 2D diffusion model $\boldsymbol{\epsilon}_{PT}(t)$ and that of the fine-tuned 2D diffusion model $\boldsymbol{\epsilon}_{FT}(t)$. We can see that the curve of $e_{FT}(t)$ is positioned under that of $e_{PT}(t)$, and we can shift the timestep of $\boldsymbol{\epsilon}_{PT}(t)$ to $\boldsymbol{\epsilon}_{PT}(t+\Delta t)$ to approximate the noise prediction error of $\boldsymbol{\epsilon}_{FT}(t)$.}

    \label{fig:diffusion_free_lunch}
    \vspace{-4mm}
\end{figure*}

The model $\boldsymbol{\epsilon}(t)$ can be a pre-trained 2D diffusion model (such as Stable Diffusion~\cite{rombach2022high}). We denote by $\boldsymbol{\epsilon}_{PT}(t)$ such a model, and investigate the behaviour of its noise prediction error, denoted by $e_{PT}(t)$. In Fig.~\ref{fig:diffusion_free_lunch}, we plot the curve (\ie, the blue colored curve) of $e_{PT}(t)$ versus $t$. We use a corpus with 15 text prompts from Magic3D~\cite{poole2022dreamfusion} to draw this curve. For each prompt $y$, we generate 16 images with VSD~\cite{wang2023prolificdreamer}. Then for each image $\boldsymbol{x}$, we apply one instance of Gaussian noise $\boldsymbol{\epsilon}$ and conduct a single diffusion step with 100 distinct timesteps. The average noise reconstruction error is then calculated for these timesteps across all prompts and images. We can see from the curve of $e_{PT}(t)$ that earlier diffusion timesteps (\eg, timestep 600) will have lower noise prediction error than later timesteps (\eg, timestep 200). Such a trend holds for almost every image sample $\boldsymbol{x}$ and noise sample $\boldsymbol{\epsilon}$ because the well-trained diffusion model is frozen in our case. Since the noise prediction error declines from $T_{\mathrm{min}}$ (\ie, late diffusion timestep) to $T_{\mathrm{max}}$ (\ie, early diffusion timestep),  we can conclude that for a given timestep $t$ and a timestep shift $0 \leq \Delta t \leq T_{\mathrm{max}} - t$, the following inequality holds:
\begin{equation}
    \mathbb{E}_{\pi, t, \boldsymbol{\epsilon}} \left[ 
        \|
            \boldsymbol{\epsilon}_\phi\left(\boldsymbol{x}_{t + \Delta t} ; t + \Delta t, y^\pi\right) - \boldsymbol{\epsilon}
        \|^2_2 
    \right] \leq 
    \mathbb{E}_{\pi, t, \boldsymbol{\epsilon}} \left[ 
        \|
            \boldsymbol{\epsilon}_\phi\left(\boldsymbol{x}_t ; t, y^\pi\right) - \boldsymbol{\epsilon}
        \|^2_2
    \right],
    \label{equ:ASD_objective}
\end{equation}
which implies that \textbf{more accurate noise predictions can be achieved at earlier diffusion timesteps}. 

The above property of diffusion models has also been observed by Yang \etal~\cite{yang2023lipschitz}, who indicated that as the timestep shifts from $T_{\mathrm{max}}$ towards $T_{\mathrm{min}}$, the variance in noise prediction increases, as evidenced by the rising Lipschitz constants, which suggests an increased instability in noise prediction and larger noise prediction errors. Such a behavior can be observed in both $\boldsymbol{\epsilon}$-prediction and $\boldsymbol{v}$-prediction models, as well as in 2D and 3D diffusion models (please refer to  Sec.~\ref{sec:more_2D_prediction_erros} for details). This can be intuitively explained as follows. When $t \rightarrow T_{\mathrm{max}}$,  $\boldsymbol{x}_t = \alpha_t \boldsymbol{x} + \sigma_t \boldsymbol{\epsilon} \rightarrow \boldsymbol{\epsilon}$, then it is easier to achieve $\boldsymbol{\epsilon}_\phi\left(\boldsymbol{x}_t ; t, y^\pi\right) \approx \boldsymbol{\epsilon}$ because the model can manage to copy the input as the output.

The similarity between Eq.~\ref{equ:ASD_objective} and the fine-tuning objective of VSD in Eq.~\ref{equ:vsd_finetuneing_objective} inspires us to investigate whether simply shifting earlier the timestep could fulfill the fine-tuning requirements of VSD without modifying the pre-trained 2D diffusion network parameters. Specifically, we employ the pretrained 2D diffusion model with shifted timestep to approximate the diffusion model of rendered images in Eq.~\ref{equ:contrastive_score_models} as $\boldsymbol{\epsilon}_{\theta}\left(\boldsymbol{x}_t ; t, \pi, y\right) \triangleq \boldsymbol{\epsilon}_\phi\left(\boldsymbol{x}_{t+\Delta t} ; t+\Delta t, y^\pi\right)$, resulting in the following  Asynchronous Score Distillation (ASD) objective function:
 \begin{equation}
\nabla_\theta \mathcal{L}_{\mathrm{ASD
}}(\theta, y) \triangleq \mathbb{E}_{\pi, t, \boldsymbol{\epsilon}} \left[ \omega(t)\left(                
    \boldsymbol{\epsilon}_{\phi}\left(\boldsymbol{x}_t; t, y^{\pi}\right) - 
    \boldsymbol{\epsilon}_{\phi}\left(\boldsymbol{x}_{t+\Delta t}; t+\Delta t, y^{\pi}\right)
\right) \frac{\partial \boldsymbol{x}}{\partial \theta} \right].
\label{equ:ASD_finetuneing_objective}
\end{equation}
We can see that rather than iteratively fine-tuning the diffusion network as in VSD, ASD achieves similar goal by shifting the timestep $t$ with an interval $\Delta t$ in each step, which is much more efficient. One key variable introduced in ASD is the timestep shift $\Delta t$, which will be discussed in the next subsection.

 \subsection{The Setting of Timestep Shift $\Delta t$}
\label{sec:ASD_interval}

 Before discussing how to set the timestep shift $\Delta t$, let's plot another curve, \ie, the noise prediction error of $\boldsymbol{\epsilon}_{\theta}\left(\boldsymbol{x}_t ; t, \pi, y\right)$  w.r.t. timestep $t$. Actually, in the process of generating $\boldsymbol{x}$ with VSD, we will have the fine-tuned model $\boldsymbol{\epsilon}_{\phi^{\prime}}\left(\mathbf{x}_t ; t, \pi, y\right)$ as the by-product, which is used to represent  $\boldsymbol{\epsilon}_{\theta}\left(\boldsymbol{x}_t ; t, \pi, y\right)$ in Eq.~\ref{equ:contrastive_score_models}. Therefore, with  fixed $\boldsymbol{x}$, $\boldsymbol{\epsilon}$ and $y$,
 the noise prediction error of the fine-tuned diffusion model, denoted by $\boldsymbol{\epsilon}_{FT}(t)$, can be calculated as $e_{FT}(t)$=
 $\|\boldsymbol{\epsilon}_{\phi^{\prime}}(t) - \boldsymbol{\epsilon}\|^2_2$.  
 
 The curve of $e_{FT}(t)$ w.r.t. $t$ (\ie, the yellow curve) is plotted in Fig.~\ref{fig:diffusion_free_lunch} by using the same data as in plotting $e_{PT}(t)$. 
 We can see that the curve of $e_{FT}(t)$ is positioned under $e_{PT}(t)$ because $e_{FT}(t)$ is obtained by the fine-tuned diffusion model $\boldsymbol{\epsilon}_{FT}$. However, as mentioned in Sec.~\ref{sec:review_score_distillation}, this fine-tuning changes the weights of pre-trained diffusion model and might damage its ability in comprehending text-image pairs. Therefore, we propose to fix the pre-trained model $\boldsymbol{\epsilon}_{PT}(t)$ but shift it to $\boldsymbol{\epsilon}_{PT}(t+\Delta t)$ to approximate the desired $\boldsymbol{\epsilon}_{FT}(t)$. 
 Referring to Fig.~\ref{fig:diffusion_free_lunch}, we could shift $\boldsymbol{\epsilon}_{PT}(t)$ to an \textbf{earlier} timestep to achieve this goal. For example, at timestep $t_0$ and with a time shift $\Delta t_0 > 0$, we can use $\boldsymbol{\epsilon}_{PT}(t_0 + \Delta t_0)$ to approximate the noise prediction error of $\boldsymbol{\epsilon}_{FT}(t_0)$.

 On the other hand, the magnitude of $\Delta t$ will vary with $t$. Let's come to another timestep $t_1$ in Fig.~\ref{fig:diffusion_free_lunch}, where $t_1$ is earlier than $t_0$. Because the decreasing speeds of both $e_{PT}$  and $e_{FT}$ will be reduced with $t$ going to $T_{\mathrm{max}}$, the magnitude of $\Delta t_1$ will be increased to approximate $e_{FT}(t_1)$. In other words, the magnitude of $\Delta t$ should grow when $t$ goes from  $T_{\mathrm{min}}$ to $T_{\mathrm{max}}$.  We heuristically set this relationship as $\Delta t = \eta(t - T_{\mathrm{min}})$, where $\eta \in [0, 1]$ is a hyper-parameter that controls the length of shift range.
 Finally, it should be pointed out that the curves in Fig.~\ref{fig:diffusion_free_lunch} will vary a little for different training iterations, rendered images $\boldsymbol{x}$ and text prompts $y$. Therefore, $\Delta t$ should fall into some range $S(t)$. In practice, we set $\Delta t \sim S(t) = \mathcal{U}[0, \eta(t - T_{\mathrm{min}})]$, which follows a uniform distribution within $0$ and $\eta(t - T_{\mathrm{min}})$. The pseudo-code of ASD is summarized in \textbf{Alg. \ref{alg:asynchronous score distillation}}, which can be applied to both prompt-specific and prompt-amortized text-to-3D tasks.

\begin{figure*}[!t]
    \centering
    \includegraphics[width=\textwidth]{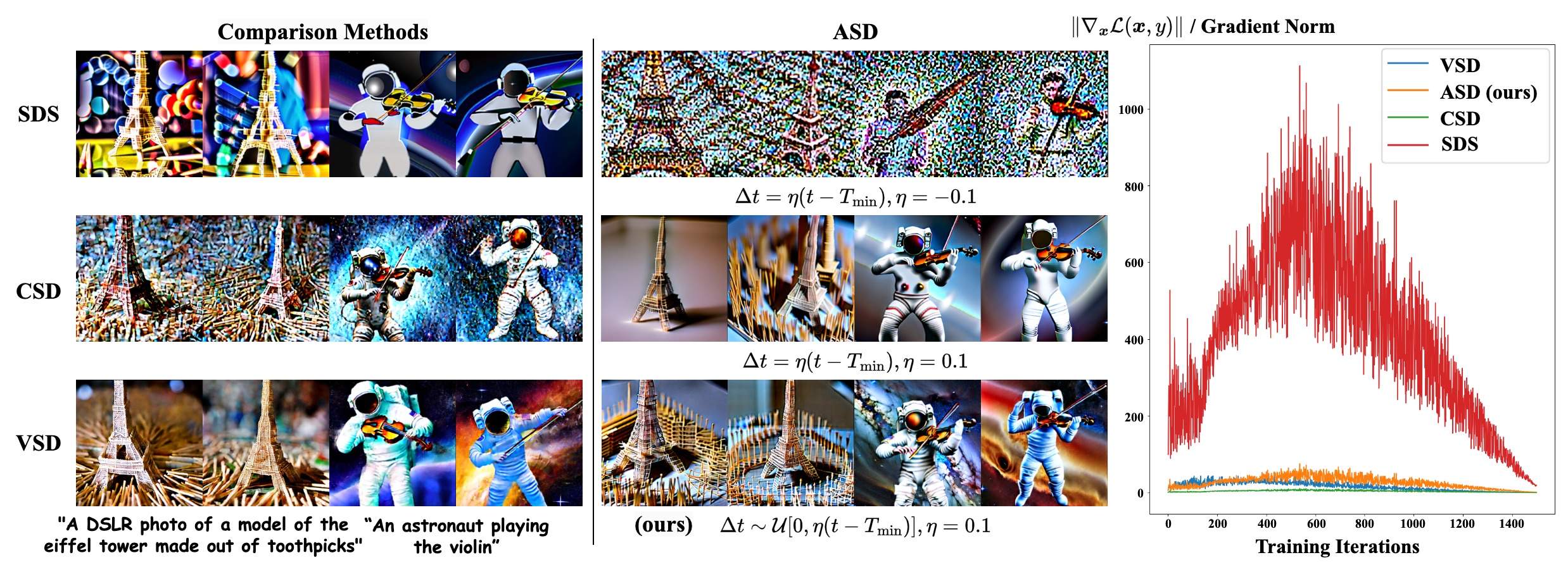}
    \caption{ \textbf{Left and middle:} 2D toy examples by SDS \cite{poole2022dreamfusion}, CSD \cite{yu2023text}, VSD \cite{wang2023prolificdreamer} and our proposed ASD. \textbf{Right:} Gradient norms generated by different methods.}
    \label{fig:2D_toy_exps}
\end{figure*}

 \textbf{2D toy experiments}. To verify the proposed timestep shift strategy, we follow the paradigm in~\cite{wang2023prolificdreamer} to test SDS, CSD, VSD and our ASD on 2D toy examples. The left column of Fig.~\ref{fig:2D_toy_exps} shows the results of SDS, CSD, VSD, and the middle column shows the results of ASD with different sampling strategies of $\Delta t$. One can see that the proposed sampling strategy $\Delta t \sim S(t)=\mathcal{U}\left[0, \eta\left(t - T_{\mathrm{min}}\right)\right]$ yields similar results to VSD~\cite{wang2023prolificdreamer}. Besides,  we show the gradient norm produced by these score distillation methods in the right column of Fig.~\ref{fig:2D_toy_exps}. One can see that the range of gradient norm produced by ASD is similar to that of VSD.  However, the gradient norm of SDS is more than 10 times larger than ASD and VSD because it needs to set CFG=100 for convergence~\cite{yu2023text,poole2022dreamfusion,wang2023prolificdreamer}. Such a large gradient may result in training instability. We append more 2D results in Sec.~\ref{sec:more_2D_experiments} to further validate our proposed sampling strategy.

\begin{algorithm}[!t]
    \caption{Asynchronous Score Distillation (ASD)}
    \label{alg:asynchronous score distillation}


    \KwIn{3D representation $\theta$; Text prompt $y$; Hyperparamter $\eta$; 2D diffusion prior $\boldsymbol{\epsilon}_{\phi}$}

    \While{not converged}{        
    
        Sample a camera pose $\pi$ 
        
        Render an image $\boldsymbol{x} = g(\theta,\pi)$

        Sample a timestep $t \sim \mathcal{U}[T_{\mathrm{min}},T_{\mathrm{max}}]$, Gaussian noise $\boldsymbol{\epsilon} \sim \mathcal{N}(0,\boldsymbol{I})$

        Sample a timestep shift $\Delta t \sim S(t)=\mathcal{U}\left[0, \eta\left(t - T_{\mathrm{min}}\right)\right]$

        $\boldsymbol{x}_t \leftarrow \alpha_t \boldsymbol{x} + \sigma_t \boldsymbol{\epsilon}$, $\boldsymbol{x}_{t + \Delta t} \leftarrow \alpha_{t+\Delta t} \boldsymbol{x} + \sigma_{t + \Delta t} \boldsymbol{\epsilon}$

        Update $\theta$ with $\Delta \theta \leftarrow \omega(t)\left(\boldsymbol{\epsilon}_\phi\left(\boldsymbol{x}_t ; t, y^\pi\right)-\boldsymbol{\epsilon}_\phi\left(\boldsymbol{x}_{t+\Delta t} ; t+\Delta t, y^\pi\right)\right) \frac{\partial \boldsymbol{x}}{\partial \theta}$

    }
 
\end{algorithm}

\begin{figure*}[!t]
    \centering
    \includegraphics[width=1\textwidth]{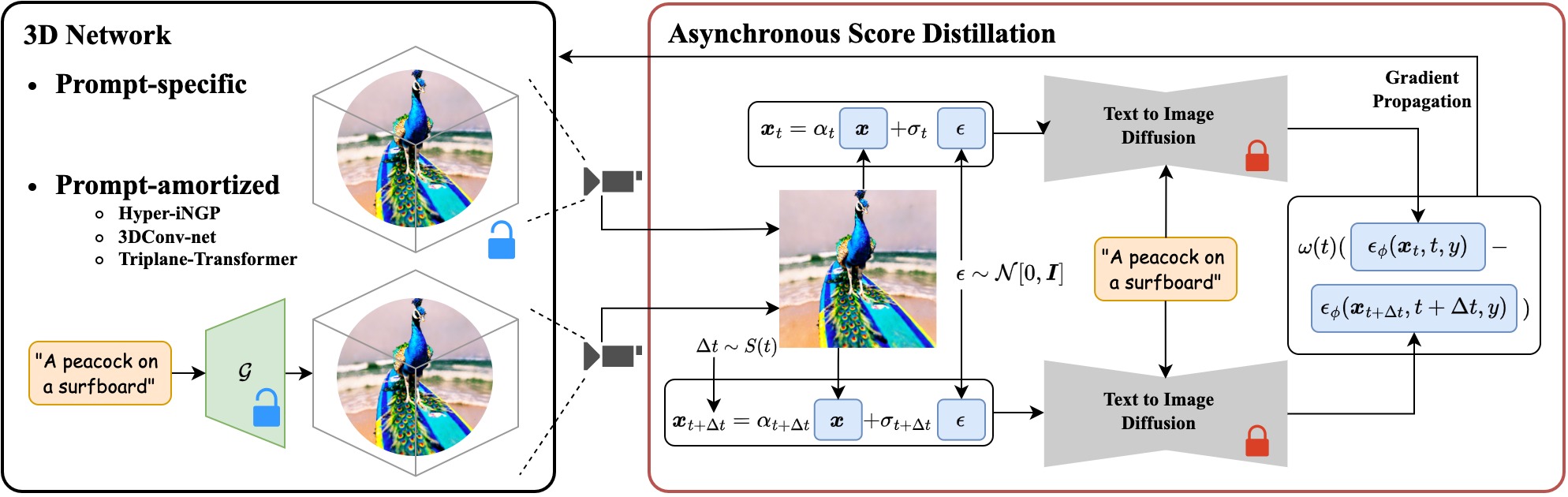}
    \caption{ Overview of Asynchronous Score Distillation (ASD). As illustrated in the left sub-figure, ASD can be employed for prompt-specific generation by optimizing 3D representations for each prompt, as well as for prompt-amortized generation by training a text-to-3D generator. The right sub-figure depicts how ASD uses the difference in noise predictions at asynchronous timesteps to update the 3D network parameters. } 
    \label{fig:overview}
\end{figure*}

 \textbf{Text-to-3D Synthesis with ASD}. As a score distillation method, ASD is open to the selection of 3D generator architectures~\cite{hong2023lrm,bahmani2023cc3d,lorraine2023att3d,muller2022instant,kerbl20233d}.  The general pipeline of ASD for text-to-3D synthesis is shown in Fig. \ref{fig:overview}. It takes a rendered image as input and diffuses it in two timesteps $t$ and $t + \Delta t$. The noise prediction difference is used as the gradient to optimize the 3D representation of generator. In this work, in addition to prompt-specific generation, as done in most existing score distillation works~\cite{poole2022dreamfusion, wang2023prolificdreamer,liang2023luciddreamer,wu2024consistent3d,ho2022classifier}, we focus more on prompt-amortized text-to-3D and conduct thorough experiments to evaluate the effectiveness of ASD with three representative architectures, \ie \textbf{Hyper-iNGP}, \textbf{3DConv-net} and \textbf{Triplane-Transformer}, using two types of 2D diffusion models, \ie \textbf{Stable Diffusion} and \textbf{MVDream}.
 
 Hyper-iNGP is adopted by ATT3D~\cite{lorraine2023att3d}, which integrates a prompt-agnostic hash-grid spatial encoding~\cite{muller2022instant} with prompt-conditioned decoding layers to output color and density.   3DConv-net~\cite{bahmani2023cc3d} is a 3D generator that maps the provided condition to voxel using 3D convolution. Triplane-Transformer is wildly adopted in 3D generation tasks~\cite{hong2023lrm,xu2024instantmesh,wei2024meshlrm,zou2023triplane,xu2024grm,xu2023dmv3d,tochilkin2024triposr,liu2023unidream,li2024m}, which facilitates 3D generation with the powerful Transformer architecture and triplane 3D representation~\cite{chan2022efficient}. We choose them in our experiments because they represent three groups of 3D generators, \ie hyper-networks~\cite{jun2023shap,babu2023hyperfields}, voxel-based network~\cite{yariv2021volume,schwarz2022voxgraf,sitzmann2019deepvoxels,tang2023volumediffusion} and triplane-based network~\cite{chan2022efficient,hong2023lrm,wei2024meshlrm,li2023instant3d,xu2024instantmesh}. All of them take CLIP~\cite{radford2021learning} text embeddings as the condition. More details of the network architectures can be found in Sec.~\ref{sec:more_3D_generator_architecture_details}. These 3D Generators can be trained with any off-the-shelf 2D diffusion model under the assistance of ASD. We choose Stable Diffusion~\cite{rombach2022high} and MVDream~\cite{shi2023mvdream} as two representative 2D diffusion models. Stable Diffusion has been widely applied in many text-to-3D works~\cite{ho2022classifier,wang2023prolificdreamer,liang2023luciddreamer,poole2022dreamfusion,lin2023magic3d,chen2023fantasia3d,tang2023dreamgaussian,yi2023gaussiandreamer}. MVDream is built on top of Stable Diffusion, and it solves the Janus problem~\cite{armandpour2023re} by producing gradient in four rendering views synchronously.


\section{Experiments}

\subsection{Experimental Settings}
\label{sec:experiment_settings}

\textbf{Comparison Methods}. We compare ASD with state-of-the-art score distillation methods, including SDS~\cite{poole2022dreamfusion}, CSD~\cite{yu2023text} and VSD~\cite{wang2023prolificdreamer}. We adhere to their official codes for training prompt-amortized text-to-3D networks. For example, the CFG~\cite{ho2022classifier} values for SDS, CSD and VSD are configured to 100, 1, and 7.5, respectively. In addition, we compare with existing prompt-amortized method ATT3D~\cite{lorraine2023att3d} (whose code is not released yet) by replicating its reported results. 

\textbf{Implementation Details.} We employ VolSDF~\cite{yariv2021volume} to render images from the 3D generators. For Stable Diffusion, we employ SD-v2.1-base~\cite{sdv2.1base} for all score distillation methods for fair comparison. As configured in VSD~\cite{wang2023prolificdreamer}, we set the CFG value as 7.5 for the pre-trained diffusion model in ASD, and 1 for the diffusion model of rendered images.
The resolution of rendered images by Hyper-iNGP is set to $256 \times 256$, while that of 3DConv-net and Triplane-Transformer is set to $64\times64$ for GPU memory considerations. Other details are in Sec.~\ref{sec:more_implementation_details}.


\textbf{Prompt Corpus}. To thoroughly evalutate the capability of ASD in prompt-amortized text-to-3D synthesis, we employ multiple datasets encompassing a range of text prompt quantities. \textbf{MG15} includes 15 prompts from Magic3D~\cite{lin2023magic3d}; \textbf{DF415} comprises 415 prompts from  DreamFusion~\cite{poole2022dreamfusion}; and \textbf{AT2520} contains 2520 compositional prompts of animals from ATT3D~\cite{lorraine2023att3d}. DL17k contains 17k compositional prompts of human with daily activities, proposed by ~\cite{li2023instant3d}. While AT2520 and DL17k provide a larger number of prompts than DF415, the prompt diversity of them is relatively low due to the predefined templates. 

To test ASD's performance with an even larger scale of prompts, we introduce a novel prompt corpus named \textbf{CP100k}. This corpus consists of 100,000 text prompts filtered from the image descriptions collected by Cap3D~\cite{luo2024scalable}, which was developed to test text-to-image model performance. 
To the best of our knowledge, it is the first time to evaluate score distillation methods on such a scale of text prompts. Meanwhile, it should be clarified that this work is focused on examining the score distillation performance rather than prompt generalization, so the test prompts share the same distribution as training prompts. 
More details of the prompt corpus are in Sec.~\ref{sec:more_corpus_details}.

\begin{figure}[!t]
\centering
\includegraphics[width=0.90\textwidth]{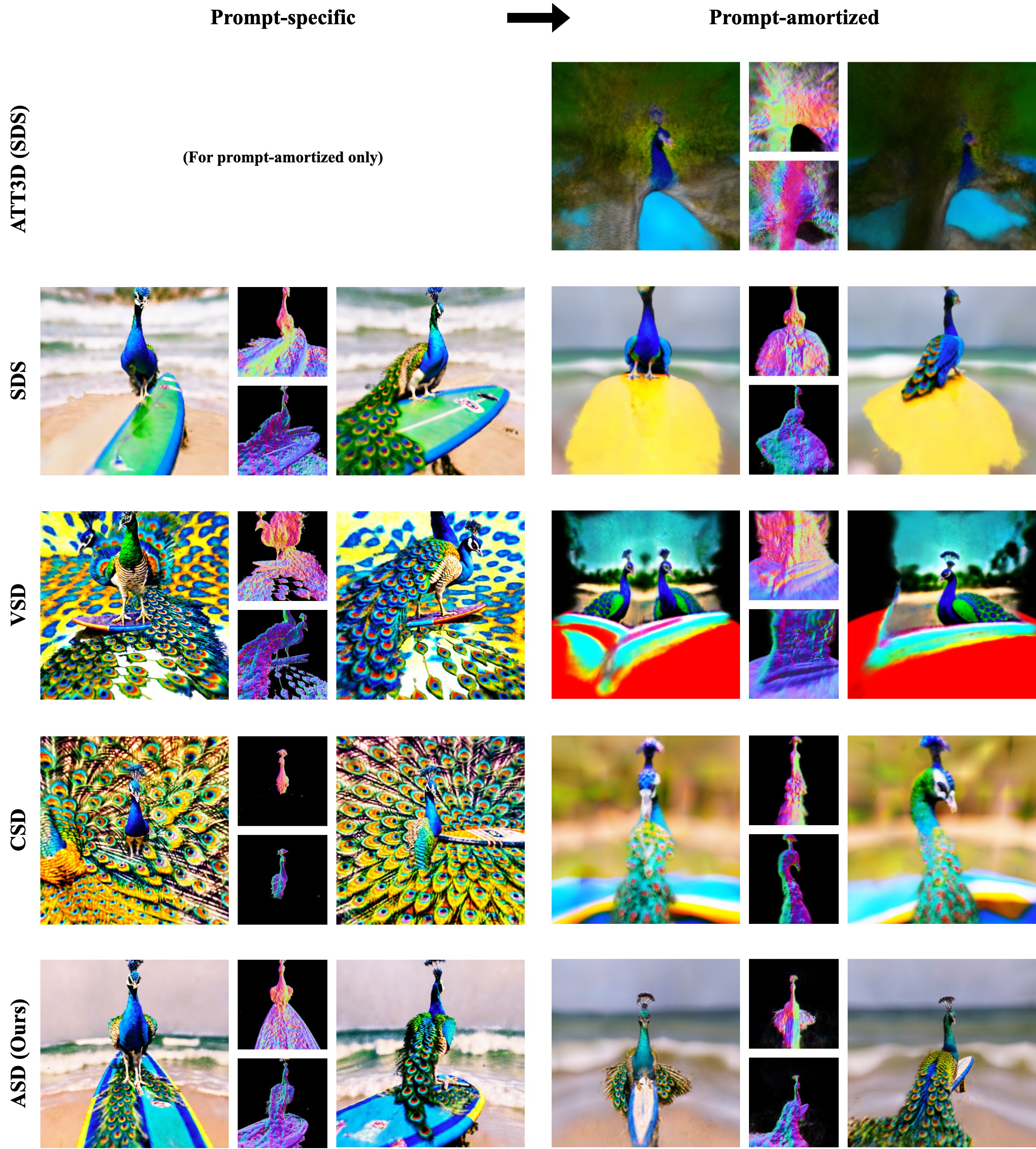}
\caption{Qualitative comparison on prompt-specific (with iNGP as the 3D representation) and prompt-amortized (with Hyper-iNGP as the 3D generator) text-to-3D results by SDS \cite{poole2022dreamfusion}, CSD \cite{yu2023text}, VSD \cite{wang2023prolificdreamer}, ATT3D \cite{lorraine2023att3d} and our ASD methods. 
}
\vspace{+1mm}
\label{fig:comparsion_in_hyper_net}
\end{figure}

\textbf{Evaluation Metrics}. We render 120 surrounding view images as the 3D synthesis result from each prompt. Similar to previous text-to-3D works~\cite{poole2022dreamfusion,lorraine2023att3d,lorraine2023att3d,li2023instant3d}, we compute the CLIP recall, \ie, the classification accuracy by applying CLIP model to the rendered images to predict the correct text prompt, as one performance metric, denoted by "R@1". Additionally, we calculate the CLIP text-image similarity between generated images and input prompts as another metric~\cite{wei2023elite,tang2023volumediffusion}, denoted by "Sim".

\subsection{Evaluation Results}

\textbf{Results with iNGP/Hyper-iNGP as 3D Representation}. The iNGP~\cite{muller2022instant} architecture is designed for prompt-specific text-to-3D generation. Hyper-iNGP has the same spatial encoding as iNGP except that the weights of the decoding layer depend on the text prompt. To eliminate the effect caused by architecture difference as much as possible, we adopt iNGP for prompt-specific  text-to-3D tasks, and Hyper-iNGP for prompt-amortized tasks. Our experiments are carried out on the MG15 dataset. For prompt-specific tasks, we optimize an individual iNGP~\cite{muller2022instant} for each MG15 prompt; while for the prompt-amortized tasks, we train a single Hyper-iNGP~\cite{lorraine2023att3d} across all MG15 prompts.  We also compare our results with ATT3D~\cite{lorraine2023att3d}, which is among the first to apply Hyper-iNGP to prompt-amortized text-to-3D tasks. ATT3D employs SDS for training and uses soft-shading~\cite{poole2022dreamfusion} (denoted as * in Tab.~\ref{tab:comparison_from_prompt-specific_to_prompt-amortized}) for rendering.

\begin{table}[!t]
\setlength{\belowcaptionskip}{-4.mm}
\setlength{\abovecaptionskip}{1.5mm}
\setlength{\tabcolsep}{0.07cm}
\centering
\begin{tabular}{@{}l|ccc|ccc@{}}
\toprule
Reference       & Method     & Sim $\uparrow$& R@1 $\uparrow$& Method           & Sim $\uparrow$& R@1 $\uparrow$\\ \midrule
ATT3D~\cite{lorraine2023att3d} & - & - & - & Hyper-iNGP* + SDS & 0.195 & 0.468                                             \\ \midrule
DreamFusion~\cite{poole2022dreamfusion}     & iNGP + SDS & 0.288   & \textbf{1.000}    & Hyper-iNGP + SDS &  0.257    &  0.918     \\
Classifier~\cite{ho2022classifier}      & iNGP + CSD & 0.280   & 0.936   & Hyper-iNGP + CSD & 0.264    & 0.972    \\
ProlificDreamer~\cite{wang2023prolificdreamer} & iNGP + VSD &  0.276 & 0.932     & Hyper-iNGP + VSD & 0.259    &    0.987  \\
Ours            & iNGP + ASD  & \textbf{0.289}    &  \textbf{1.000}   & Hyper-iNGP + ASD & \textbf{0.284}    & \textbf{1.000}    \\ \bottomrule
\end{tabular}
\caption{Quantitative comparison on prompt-specific (with iNGP as the 3D representation) and prompt-amortized (with Hyper-iNGP as the 3D generator) text-to-3D results by SDS \cite{poole2022dreamfusion}, CSD \cite{yu2023text}, VSD \cite{wang2023prolificdreamer}, ATT3D \cite{lorraine2023att3d} and our ASD methods.}
\label{tab:comparison_from_prompt-specific_to_prompt-amortized}
\end{table}

\begin{figure}[!t]
    \setlength{\belowcaptionskip}{-2.mm}
    \setlength{\abovecaptionskip}{-0.3mm}
    \centering
    \includegraphics[width=1\textwidth]{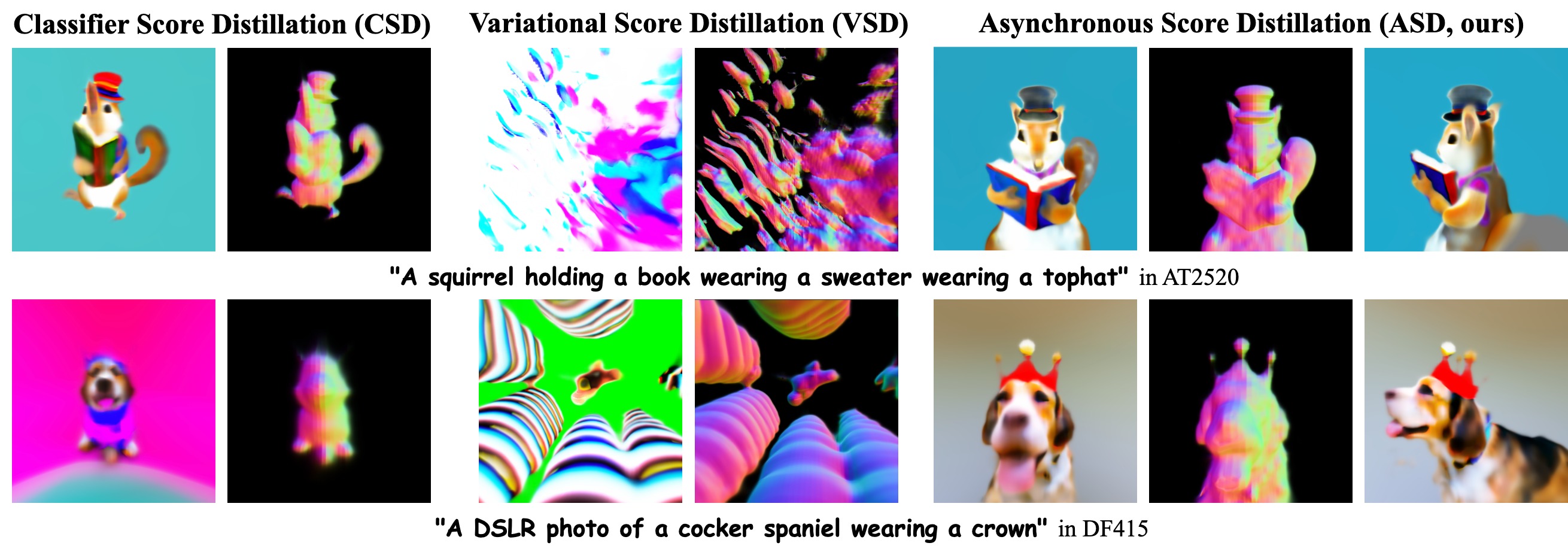}
    \caption{Qualitative comparison among CSD \cite{yu2023text}, VSD \cite{wang2023prolificdreamer} and our ASD  (with 3DConv-net as generator) on AT2520 and DF415 corpuses. SDS is not compared because it encounters numerical instability in this experiment.}
    \label{fig:score_distillation_3DConv_net_comparison}
\end{figure}


The qualitative and quantitative results are shown in Fig.~\ref{fig:comparsion_in_hyper_net} and Tab.~\ref{tab:comparison_from_prompt-specific_to_prompt-amortized}, respectively. We can see that the existing methods suffer from performance decrease when transiting from prompt-specific to prompt-amortized tasks, as evidenced by the decreased CLIP similarity and recall in Tab.~\ref{tab:comparison_from_prompt-specific_to_prompt-amortized}. It is worth mentioning that training Hyper-net with SDS requires turning on the spectral normalization~\cite{miyato2018spectral} in the linear layers, otherwise the training will fail due to numerical instability. This observation is consistent with what reported in ATT3D~\cite{lorraine2023att3d}. This is because SDS suffers from large gradient norm (please also refer to Fig.~\ref{fig:2D_toy_exps} and the discussions therein), which makes Hyper-iNGP hard to converge. As can be seen in Fig.~\ref{fig:comparsion_in_hyper_net}, ATT3D results in wrong geometry by using soft shading and SDS for training. For CSD, we see that it fails to optimize the full geometry, as shown by the shrunk peacock in both prompt-amortized and prompt-amortized results. For VSD, it tends to generate content drifts~\cite{shi2023mvdream}, resulting in repetitive patterns and abnormal geometry. It may fail to generate reasonable contents in both prompt-specific and prompt-amortized tasks. In contrast, our proposed ASD works very stable across the two tasks, yielding not only outstanding quantitative scores but also high quality 3D contents.

\begin{table}[!t]
\setlength{\belowcaptionskip}{-2.mm}
\setlength{\abovecaptionskip}{1.5mm}
\setlength{\tabcolsep}{0.22cm}
\centering
\begin{tabular}{@{}ccccccc@{}}
\toprule
\multirow{2}{*}{Method} & \multicolumn{2}{c}{DF415} & \multicolumn{2}{c}{AT2520} & \multicolumn{2}{c}{CP100k} \\ \cmidrule(l){2-7} 
                        & Sim $\uparrow$        & R@1 $\uparrow$        & Sim $\uparrow$         & R@1 $\uparrow$        & Sim $\uparrow$         & R@1 $\uparrow$        \\ \midrule
SDS                     & $\times$           & $\times$           & $\times$            & $\times$           & $\times$            & $\times$           \\
CSD                     &  0.176            & 0.062            & 0.279        & 0.037       &  0.195       & 0.108       \\
VSD                     & 0.158       & 0.002       & 0.115        & 0.001       & 0.103       & 0.000             \\
ASD (ours)              & \textbf{0.237}       & \textbf{0.276}       & \textbf{0.285}        & \textbf{0.058}       &   \textbf{0.199}      &  \textbf{0.117}      \\ \bottomrule
\end{tabular}
\caption{Quantitative comparison on prompt-amortized text-to-3D with 3DConv-net as generator. Symbol $\times$ denotes that the training fails due to numerical instability. }
\label{tab:comparison_3DConv-net}
\end{table}

\begin{figure}[!t]
    \setlength{\belowcaptionskip}{-6.mm}
    \setlength{\abovecaptionskip}{-0.1mm}
    \centering
    \includegraphics[width=1.0\textwidth]{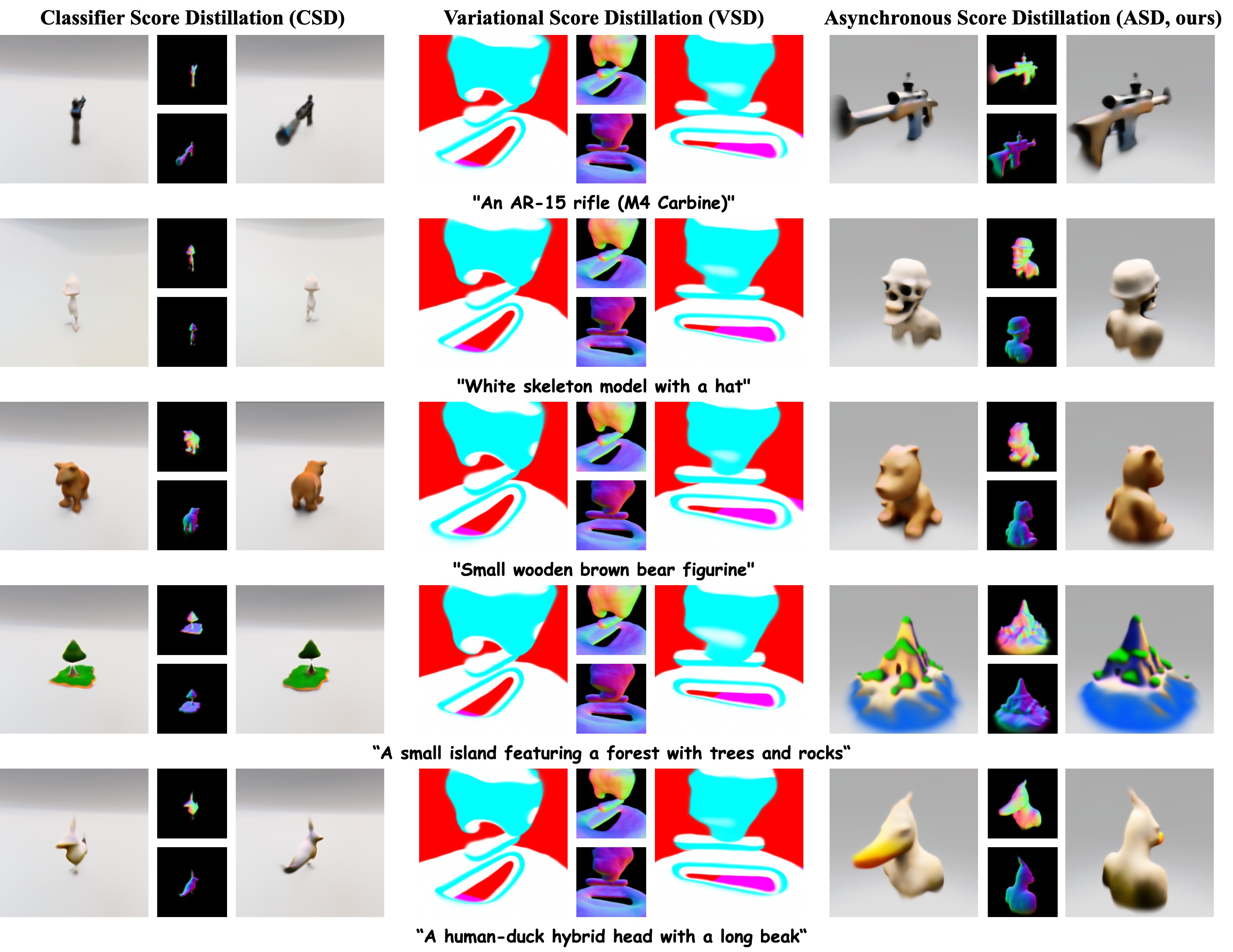}
    \caption{The scalability comparison with CSD~\cite{yu2023text} and VSD~\cite{wang2023prolificdreamer} on CP100k corpus.}
    \label{fig:scalability_comparison}
\end{figure}

\textbf{Results with 3DConv-net as 3D Generator}. The issues of existing score distillation methods  either persist or become more pronounced when replacing Hyper-iNGP to 3DConv-net as the 3D generator. We find that training SDS with 3DConv-net always fails within several thousand iterations, even using spectral or other normalization techniques. This issue stems from that deeper network is more sensitive to large gradients~\cite{he2016deep} caused by SDS. Therefore, we only compare the results of other methods in Fig.~\ref{fig:score_distillation_3DConv_net_comparison}.  We see that CSD outputs acceptable results on AT2520, but its results on DF415, which has more varied prompts, are consistently smaller than anticipated. Such a phenomenon has been observed when Hyper-iNGP is used as the generator, which underlines CSD's inability to reliably guide the 3D generator to produce geometries aligned with the text prompts. As for VSD, it leads to rather abnormal results, failing to match the text prompts. This can be attributed to its fine-tuning of the pre-trained 2D diffusion model, which severely compromises VSD's text-image comprehending ability. In comparison, our proposed ASD, with 3DConv-net as the generator, yields improved outcomes, as evidenced by the visual results in Fig.~\ref{fig:score_distillation_3DConv_net_comparison} and the enhanced metric scores in Tab.~\ref{tab:comparison_3DConv-net}. 


\textbf{Scalability}. In this section, we evaluate the scalability of competing methods by using as many as 100k prompts in the CP100k dataset with 3DConv-net as the generator. The results are shown in Fig.~\ref{fig:scalability_comparison} and Tab.~\ref{tab:comparison_3DConv-net}. Due to the issue of numerical instability, SDS is not involved  in this experiment. We can see that the outcomes of CSD are significantly diminished with uniformly small-sized shapes across all prompts. There is also a lack of variety since most outputs exhibit similar patterns. The  results of VSD are also degenerated, displaying almost identical and anomalous outcomes for the text prompts. This resembles the phenomenon of mode collapse often encountered in bi-level optimization \cite{thanh2020catastrophic}, which also highlights the importance of fixing the 2D diffusion model when training with such a large number of text prompts. In comparison, ASD is able to produce much higher quality outcomes across the text prompts, showcasing its capability in large-scale training with numerous text prompts as inputs.

\subsection{Ablation Study}

\begin{figure}[!t]

\includegraphics[width=1.0\textwidth]{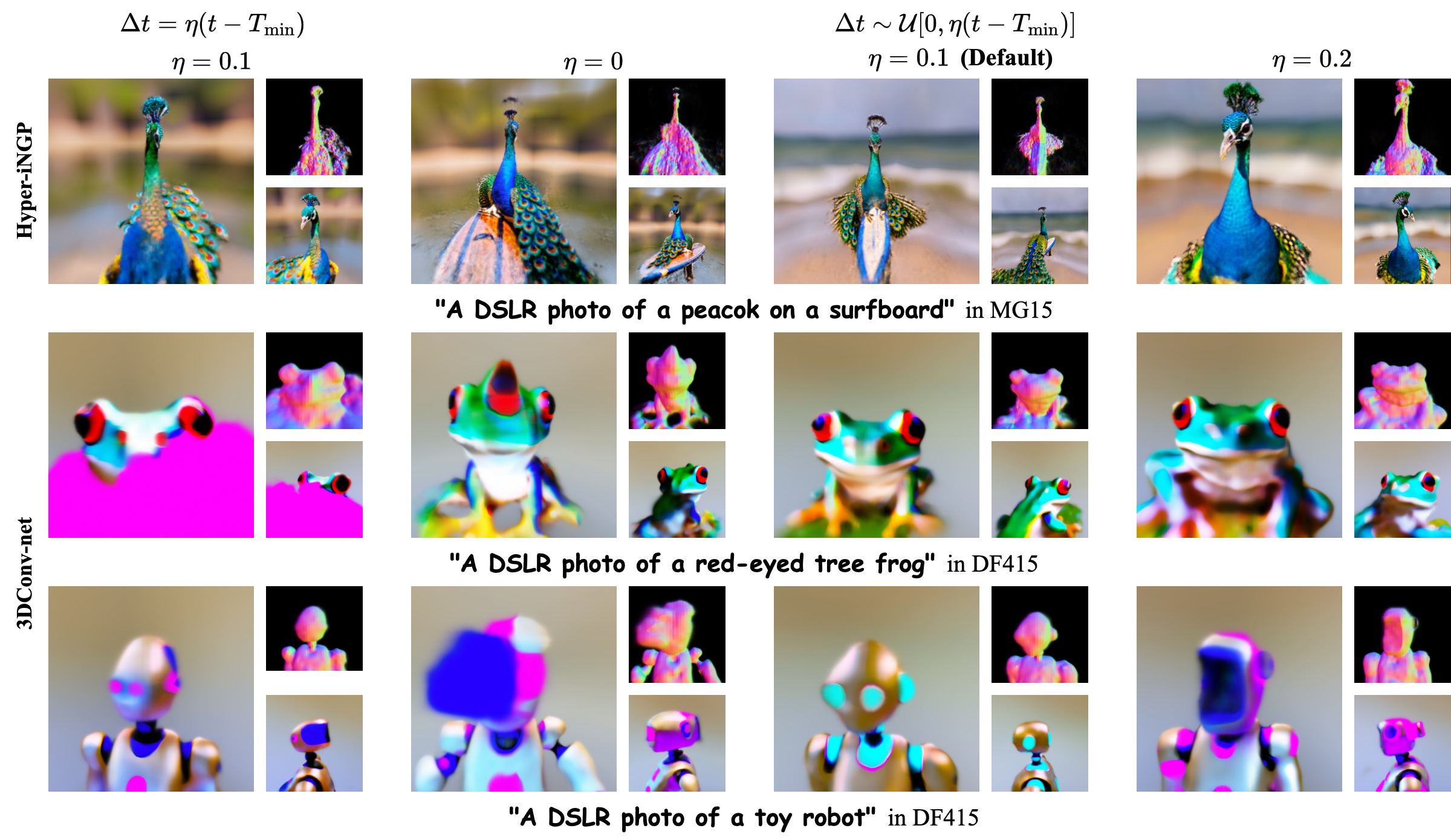}
\caption{The qualitative results of the ablation study on the timestep interval $\Delta t$.}
\label{fig:ablation_study}
\end{figure}

\begin{table}[!t]
\setlength{\belowcaptionskip}{-4.mm}
\setlength{\abovecaptionskip}{1.5mm}
\setlength{\tabcolsep}{0.22cm}
\centering
\begin{tabular}{@{}lccc@{}}
\toprule
  & Param   & Sim $\uparrow$ & R@1 $\uparrow$ \\ \midrule
\multirow{2}{*}{$\Delta t = \eta(t - T_{\mathrm{min}})$}                 & $\eta = 0.1$  & 0.214                  & 0.178                 \\
                                         & $\eta = 0.2$             &  0.214                   &  0.180                 \\
\multirow{3}{*}{$\Delta t \sim \mathcal{U}[0, \eta(t - T_{\mathrm{min}})]$}                 & $\eta = 0$     &  0.235                  &  0.267                 \\
                                         & $\eta = 0.1$      &  \textbf{0.237}                  &  \textbf{0.276}                 \\
                                         & $\eta = 0.2$                  &  0.229                  & 0.237                  \\ \bottomrule
\end{tabular}
\caption{The quantitative results of the ablation study on the timestep interval $\Delta t$.}
\label{tab:ablation_sd_diffusion_prior}
\end{table}

In this section, we perform ablation studies to evaluate the settings of timestep shift $\Delta t \sim S(t)=\mathcal{U}\left[0, \eta\left(t - T_{\mathrm{min}}\right)\right]$ from several aspects. The qualitative and quantitative results are shown in Fig.~\ref{fig:ablation_study} and Tab.~\ref{tab:ablation_sd_diffusion_prior}, respectively. 

\textbf{Importance of Timestep Shift}. We use $\eta = 0$ (\ie, no timestep shift) as a baseline to evaluate the necessity of introducing timestep shift $\Delta t$. From Fig.~\ref{fig:ablation_study} and Tab.~\ref{tab:ablation_sd_diffusion_prior}, we see that while it can generate plausible results, the model is prone to generating shapes that do not make sense, such as the so-called Janus problem~\cite{armandpour2023re}. Examples include a frog with an extra eye, robot face with block-like features, and a peacock with tails at both the front and back.  This is because the non-shifted diffusion model will align more with the 2D image distribution, tending to generate redundant contents and unreasonable geometry along the training. By introducing a timestep shift, our proposed ASD demonstrates advantages in achieving more coherent and visually pleasing results.
 

\textbf{Range of Timestep Shift}. By setting $\eta = 0.2$, we allow $\Delta t$ to be sampled from a large range. However, this might not be a good choice. In the extreme case, for any timestep $t$  we can set a large interval $\Delta t$ such that $t + \Delta t = T_{\mathrm{max}}$, then the noise prediction becomes $\boldsymbol{\epsilon}_{\phi}(\boldsymbol{x}_t; t, y^{\pi}) \approx \boldsymbol{\epsilon}$, so that ASD is degraded to SDS, which cannot perform well under CFG=7.5~\cite{poole2022dreamfusion}. In practice, we find a larger $\eta$ tends to result 3D contents with larger size and rounded shapes, \eg, the peacock with closer views, or the frog with larger size, as shown in Fig.~\ref{fig:ablation_study}. Therefore, we set $\eta = 0.1$ in all our experiments. 

\textbf{Deterministic or Random Shift}. If we set $\Delta t = \eta\left(t-T_{\min }\right)$, it assumes that the diffusion model of rendered images can be approximated by the pre-trained one with a fixed and deterministic timestep shift. As shown in Fig.~\ref{fig:ablation_study} and Tab.~\ref{tab:ablation_sd_diffusion_prior}, it reduces the chance to generate correct geometry and colors. Randomly sampling $\Delta t$ in a range is more effective, which is adopted in our method.

\begin{figure}[!t]
\includegraphics[width=\textwidth]{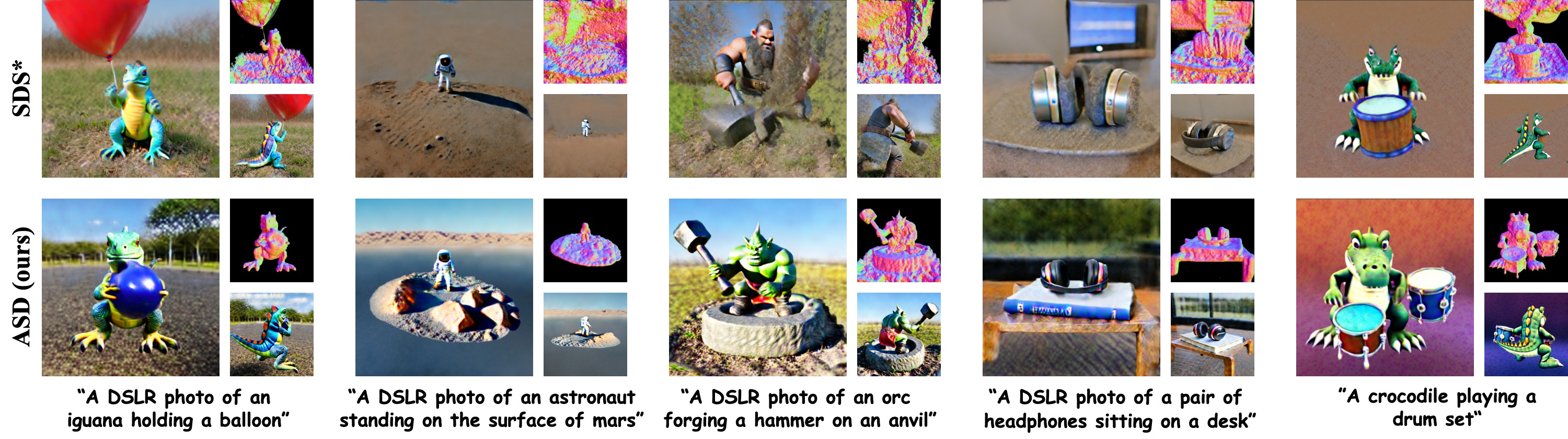}
\caption{Qualitative comparison between SDS* and ASD on prompt-specific text-to-3D generation, with iNGP as 3D representation and  MVDream as 2D diffusion prior.}
\label{fig:mvdream_specific_compare}
\end{figure}

\subsection{Results with MVDream}

As a score distillation method, ASD is open to the choice of 2D diffusion models. In this section, we evaluate ASD's compatibility with another representative 2D diffusion model, MVDream \cite{shi2023mvdream}. To conduct score distillation, MVDream takes four views as input for rendering, and  explicitly uses the camera poses as prompts. We conduct comparison and ablation study in prompt-specific optimization with iNGP as the 3D representation, as well as prompt-amortized text-to-3D with Triplane-Transformer as the 3D generator.

\textbf{Results with iNGP as 3D Representation}. MVDream officially implements a modified SDS method by incorporating the CFG re-scale technique~\cite{lin2024common} to alleviate large gradient norms caused by SDS. We refer to this modified SDS as SDS*. We qualitatively compare the performance of SDS* and ASD on prompt-specific text-to-3D. The results are shown in Fig.~\ref{fig:mvdream_specific_compare}. It can be seen that SDS* produces abnormal geometry with solid matter covering most of the 3D space, and it generates grayish textures. In contrast, ASD generates more natural geometry and textures. More results of ASD can be found in Fig.~\ref{fig:teaser}.

\begin{figure}[!t]
\includegraphics[width=1.0\textwidth]{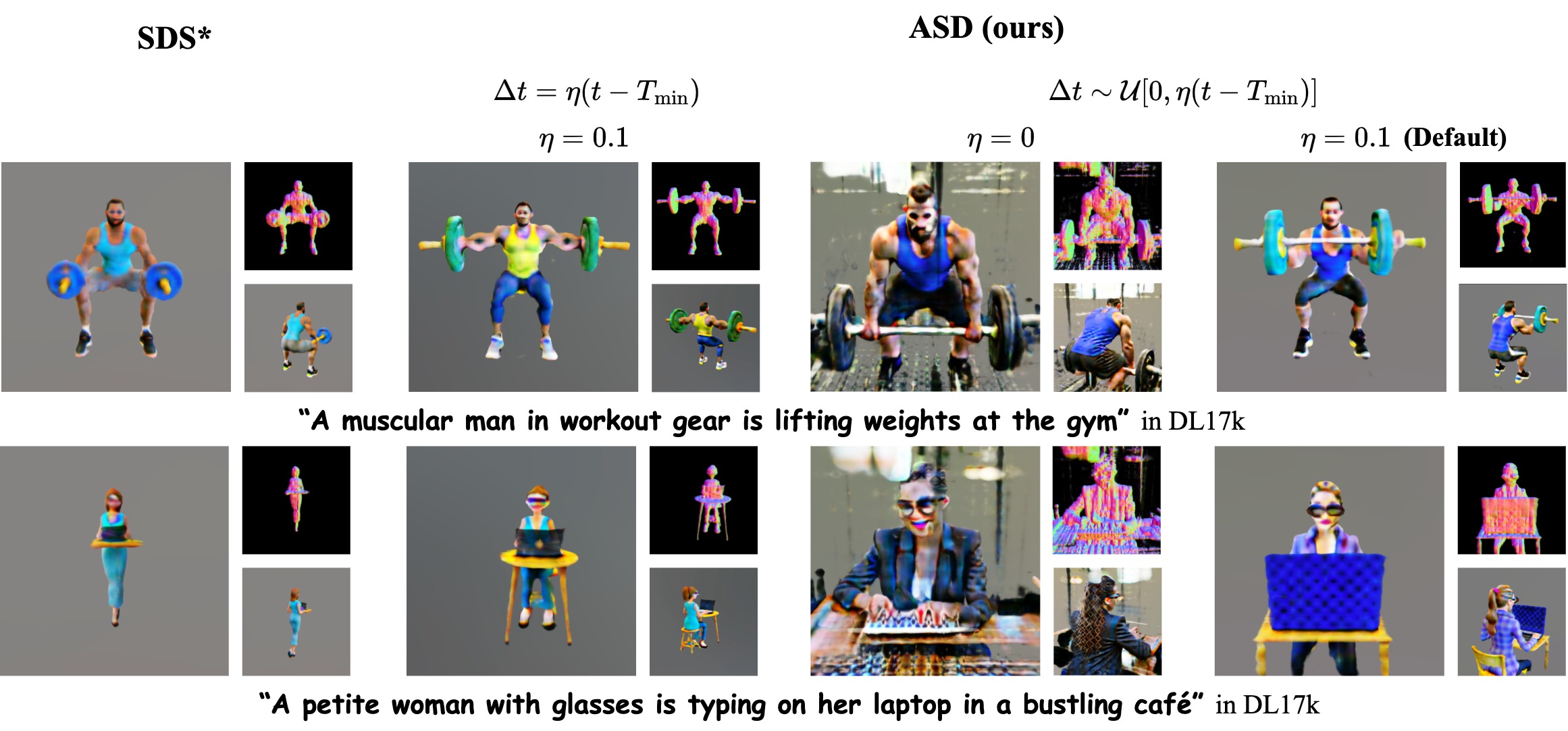}
\caption{Qualitative comparison among SDS*~\cite{shi2023mvdream} and our ASD on DL17k corpus with Triplane-Transformer as 3D generator and MVDream as 2D diffusion prior.}
\label{fig:mvdream_triplane_compare}
\end{figure}

\textbf{Results with Triplane-Transformer as 3D Generator}. We then employ MVDream for prompt-amortized text-to-3D by using Triplane-Transformer as the 3D generator. In addition to the comparison with SDS*, we ablate ASD without timestep shift to further solidify our proposed asynchronous timesteps. The experiments are conducted on DL17k corpus. As shown in Fig.~\ref{fig:mvdream_triplane_compare}, SDS* tends to produce small geometries. By using ASD with a deterministic timestep shift, \ie $\Delta t=\eta\left(t-T_{\min }\right)$, the results are improved yet still unsatisfactory. Without any timestep shift in ASD, \ie, $\eta = 0$, the 3D results have some floating patterns. This happens because without a timestep shift, the model fails to align the distribution of rendered images with the prior distribution of pre-trained diffusion model. By using a random timestep shift $\Delta t \sim \mathcal{U}\left[0, \eta\left(t-T_{\min }\right)\right]$ and the magnitude of $\eta = 0.1$ in ASD, the results are significantly improved, which is also reflected in the metrics shown in Tab.~\ref{tab:ablation_comparison_mv_diffusion_prior}.

\begin{table}[!t]
\centering
\setlength{\belowcaptionskip}{-4.mm}
\setlength{\abovecaptionskip}{1.5mm}
\setlength{\tabcolsep}{0.3cm}
\begin{tabular}{lllll}
\toprule
                     & & & Sim$\uparrow$ & R@1$\uparrow$ \\ \midrule
                SDS* & & & 0.200 & 0.159 \\
\multirow{3}{*}{ASD} & {$\Delta t = \eta(t - T_{\mathrm{min}})$}& $\eta=0.1$ & 0.205 & 0.231 \\
                     & {\multirow{2}{*}{$\Delta t \sim \mathcal{U}\left[0, \eta(t - T_{\mathrm{min}})\right]$}} & {$\eta=0$} & {0.213} & {0.293} \\
                                                & & $\eta=0.1$ & \textbf{0.219} & \textbf{0.294} \\ \bottomrule
\end{tabular}
\caption{Comparison with SDS* and ablation study on ASD using MVDream as the 2D diffusion model.}
\label{tab:ablation_comparison_mv_diffusion_prior}
\end{table}

\subsection{Discussions with Data-Driven Methods}

\begin{figure}[!t]
\centering
\setlength{\belowcaptionskip}{-2.mm}
\setlength{\abovecaptionskip}{-0.3mm}
\includegraphics[width=1.0\textwidth]{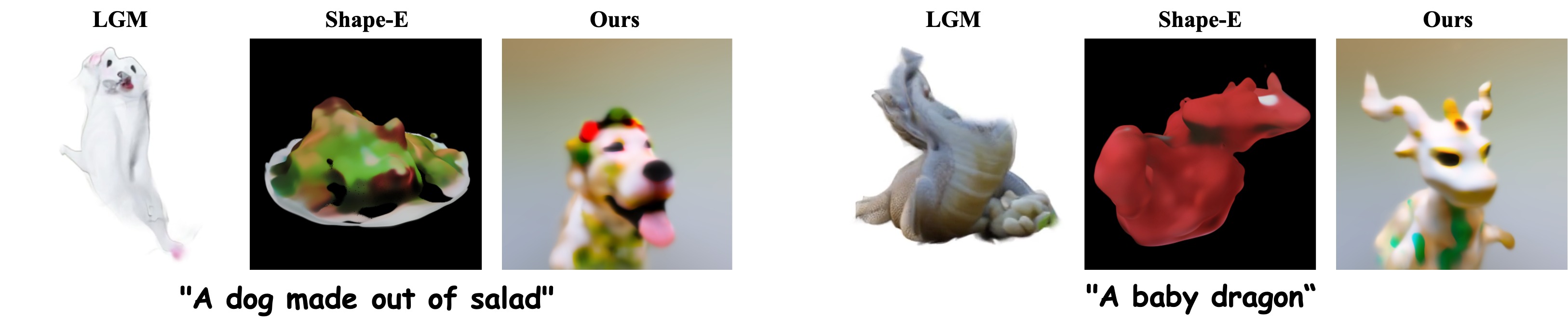}
\vspace{+2mm}
\caption{The visual comparison with data-driven methods LGM~\cite{tang2024lgm} and Shape-E~\cite{jun2023shap}.}
\label{fig:comparison_data_driven}
\end{figure}

Our proposed method differs from existing data-driven methods~\cite{hong20243dtopia,zhang2024gaussiancube,tang2023volumediffusion,tang2024lgm,jun2023shap} in that we do not require any 3D dataset to train the 3D generator. If the test text prompts fall into the training distribution, these supervised data-driven methods may generate better quality outputs than our unsupervised method. However, by leveraging the strong prior information in pre-trained 2D diffusion models, our method has better generalization capability to the test prompts. By using our 3DConv-net trained on DF415 corpus as an example, we compare our results with open-sourced data-driven 3D generators LGM~\cite{tang2024lgm} and Shape-E~\cite{jun2023shap}. Fig.~\ref{fig:comparison_data_driven} shows the qualitative comparison on some text prompt inputs, which are are out of the training distribution. We can see that LGM and Shape-E output poor results. In contrast, ASD can still work well by exploiting the powerful diffusion priors in pre-trained 2D models.

\section{Conclusion and Limitations}

In this paper, we presented Asynchronous Score Distillation (ASD), a novel score distillation method that can assist 2D diffusion prior in training 3D generators with a scalable size of text prompts. By shifting the diffusion timestep to earlier stages, our ASD can effectively predict the noise prediction error to align the diffusion model with the distribution of rendered images, while preserving the superior text comprehension capability of pre-trained models, thus facilitating stable training with high-fidelity generation results. Our extensive experiments revealed that ASD performed consistently well on datasets of various sizes, being able to manage as much as 100k prompts.

Though ASD has shown improvements over earlier score distillation approaches, there remain some limitations.
For man-made objects that have very regular shapes, such as chairs or airplanes, the performance of our model will lag behind those data-driven methods, which benefit from an abundance of relevant data. 
We foresee opportunities to combine the advantages of data-driven and score distillation methodologies to improve text-to-3D capabilities in a more comprehensive manner in the future research.

\section{Acknowledgement}
This work is supported in part by the Beijing Science and Technology Plan Project Z231100005923033, and the InnoHK program.

\clearpage
%
%
\bibliographystyle{splncs04}
\bibliography{main}

\clearpage

\appendix
\section*{\centering{Appendix}}

\renewcommand\thesection{A.\arabic{section}}

\noindent In this appendix, we provide the following materials:
\begin{itemize}

\item Sec.~\ref{sec:more_2D_prediction_erros}: more illustrations of noise prediction error $\boldsymbol{\epsilon}_{F T}(t)$ by different diffusion models $\boldsymbol{\epsilon}(t)$ (referring to Sec.~\ref{se:objective_of_asd} and Fig.~\ref{fig:diffusion_free_lunch} in the main paper);
\item Sec.~\ref{sec:more_2D_experiments}: more 2D toy experiments of different methods (referring to Sec.~\ref{sec:ASD_interval} and Fig.~\ref{fig:2D_toy_exps} in the main paper);
\item Sec.~\ref{sec:more_3D_generator_architecture_details}: more details of 3D generator architectures (referring to Sec.~\ref{sec:ASD_interval} and Fig.~\ref{fig:overview} in the main paper);
\item Sec.~\ref{sec:more_corpus_details}: more corpus details (referring to Sec.~\ref{sec:experiment_settings} in the main paper);
\item Sec.~\ref{sec:more_implementation_details}: more implementation details (referring to Sec.~\ref{sec:experiment_settings} in the main paper);
\end{itemize}


\section{More Illustrations of Noise Prediction Error}
\label{sec:noise_prediction_behavior}
\begin{figure}[!t]
    \centering
    \includegraphics[width=\textwidth]{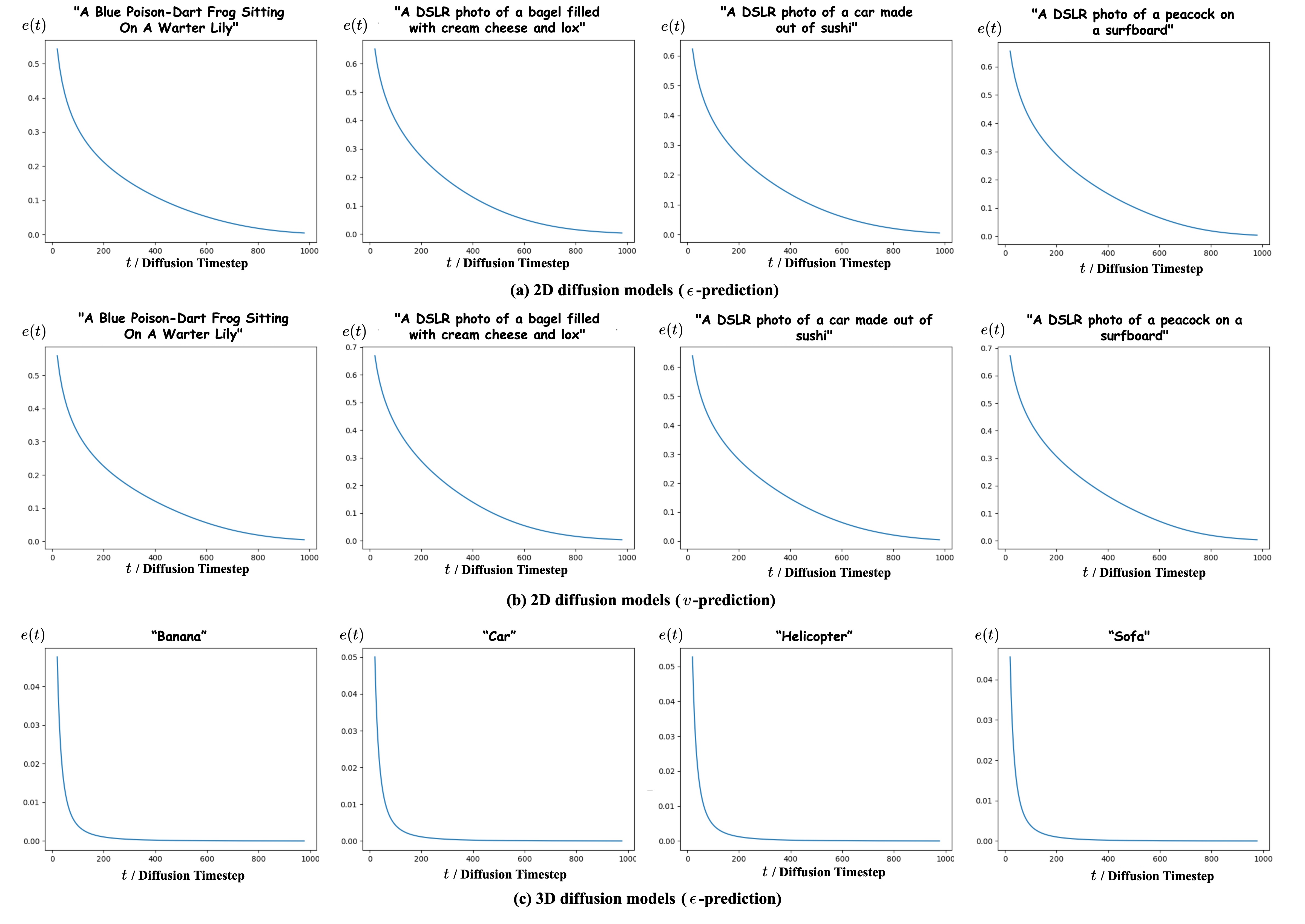}
    \caption{The behavior of noise prediction error of different diffusion models, including (a) 2D $\boldsymbol{\epsilon}$-prediction~\cite{sdv2.1base}  diffusion model, (b) 2D $v$-prediction~\cite{sdv2.1} diffusion model, and (c) 3D diffusion model. Zoom in for a better view.}
    \label{fig:noise_prediciton_supp}
\end{figure}

In this section, we provide more illustrations of the noise prediction error by various pre-trained diffusion models, including the 2D $\boldsymbol{\epsilon}$-prediction model~\cite{rombach2022high,sdv2.1base} and the $\boldsymbol{v}$-prediction model \cite{salimans2022progressive, sdv2.1}, and the 3D diffusion model \cite{cao2023large
}. We plot the  the noise prediction error against timesteps in Fig.~\ref{fig:noise_prediciton_supp}.  For each text prompt displayed at the top of the sub-figures, we use it as the condition to generate 16 samples. We then introduce a single instance of Gaussian noise to each sample and execute one diffusion step at 100 different timesteps. The DDPM~\cite{ho2020denoising} is used as the noise scheduler, as done in VSD~\cite{wang2023prolificdreamer}. The average noise reconstruction error is then calculated over the timesteps and the 16 data samples.

\label{sec:more_2D_prediction_erros}

\textbf{2D $\boldsymbol{\epsilon}$-prediction diffusion model}. The $\boldsymbol{\epsilon}$-prediction model is widely adopted in the field of text-to-3D synthesis~\cite{wang2023prolificdreamer,liang2023luciddreamer,wu2024consistent3d,shi2023mvdream,qiu2023richdreamer}. In our tests, we employ the commonly used SD-v2.1-base model~\cite{sdv2.1base}. The noise prediction error curves for four prompts sourced from Magic3D~\cite{lin2023magic3d} are presented in Fig.~\ref{fig:noise_prediciton_supp}(a), from which we see a clear decrease of noise prediction error with the timestep going from $T_{\mathrm{min}}$ to $T_{\mathrm{max}}$.

\textbf{2D $\boldsymbol{v}$-prediction diffusion model}. The $\boldsymbol{v}$-prediction model, introduced by Salimans \etal~\cite{salimans2022progressive}, accelerates the generation process by predicting velocity rather than noise. We test this model using the well-known SD-v2.1\cite{sdv2.1} with 4 prompts sourced from Magic3D~\cite{lin2023magic3d}. To calculate the noise prediction error, we convert the velocity predictions into noise predictions~\cite{salimans2022progressive}. As depicted in Fig.~\ref{fig:noise_prediciton_supp}(b), the $\boldsymbol{v}$-prediction model also exhibits reduced prediction errors as the timestep goes from $T_{\mathrm{min}}$ to $T_{\mathrm{max}}$.

\textbf{3D diffusion model}. Apart from the above 2D diffusion models, we also conduct experiments on a 3D diffusion model DiffTF~\cite{cao2023large}, which is a 3D generator trained on 3D object datasets~\cite{wu2023omniobject3d}. It is configured with $\boldsymbol{\epsilon}$-prediction and performs the diffusion process on tri-plane~\cite{chan2022efficient}. As shown in Fig.~\ref{fig:noise_prediciton_supp}(c), its noise prediction error $e(t)$ also reduces as timestep $t$ increases, which is similar to 2D diffusion models. In particular, $e(t)$ drops rapidly before $t = 200$. This is mainly caused by the much smaller scale (\eg, 6k 3D objects) of the 3D dataset~\cite{deitke2023objaverse} compared with the 2D datasets~\cite{schuhmann2022laion} (\eg, 2B text-image pairs). Therefore, the network tends to overfit the 3D data with smaller prediction error.

\begin{figure}[!t]
\includegraphics[width=\textwidth]{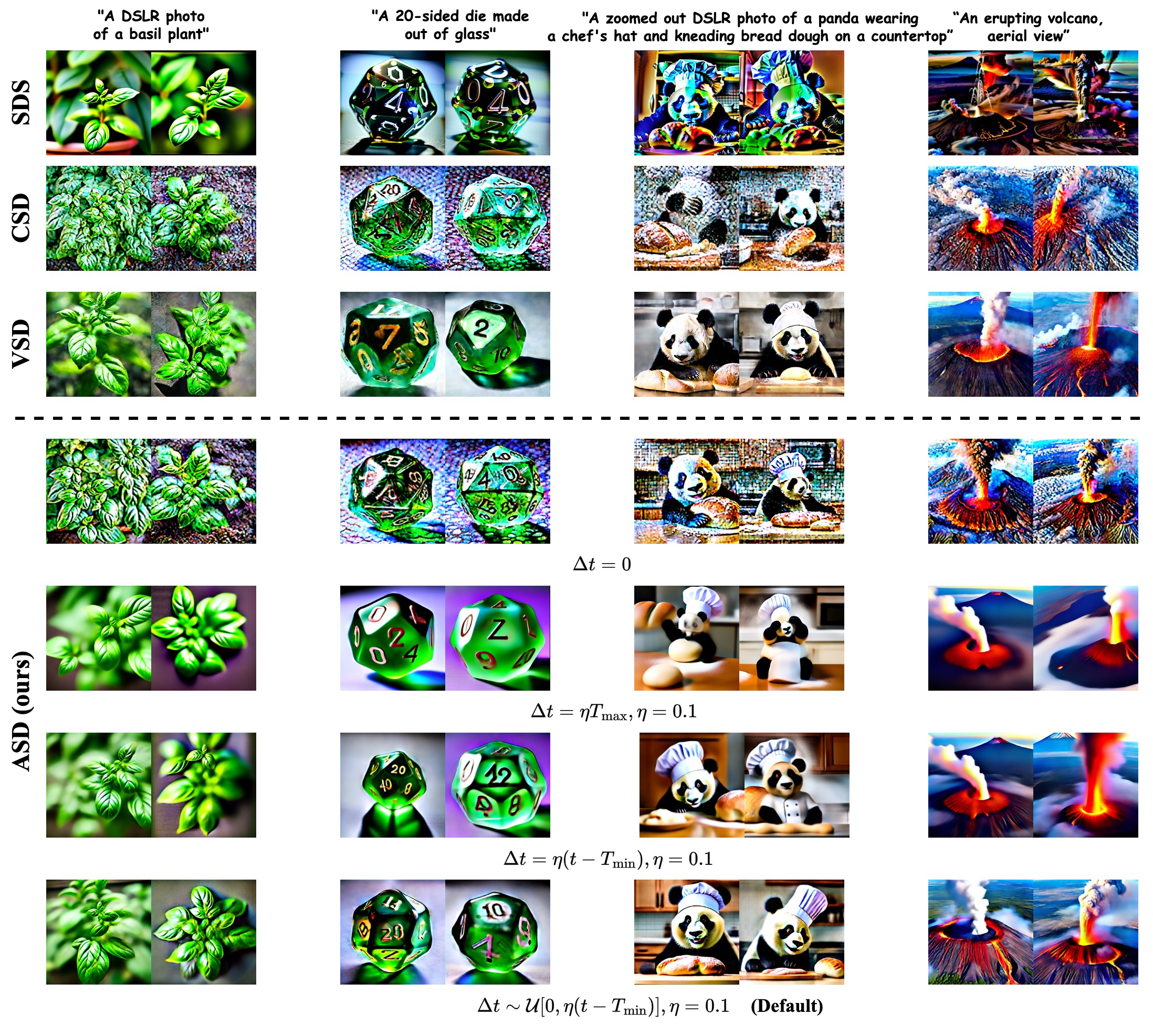}
\caption{2D toy experiments by SDS~\cite{poole2022dreamfusion}, CSD~\cite{ho2022classifier}, VSD~\cite{wang2023prolificdreamer} and our ASD with different settings of $\Delta t$. }
\label{fig:2D_toy_exps_supp}
\end{figure}

\section{More 2D Toy Experiments}
\label{sec:more_2D_experiments}

To further validate the effectiveness of the introduced timestep interval $\Delta t$ in our ASD, we provide more 2D toy experiments in Fig.~\ref{fig:2D_toy_exps_supp}, covering a wild range of subjects, \ie, plants, objects, animals, and scenes. 

From Fig.~\ref{fig:2D_toy_exps_supp}, we can see that SDS~\cite{poole2022dreamfusion} and CSD~\cite{yu2023text} do not perform very well. SDS generates high-saturation results because of the large CFG~\cite{ho2022classifier}, while CSD shows noisy and blurred patterns so that the subjects are difficult to identify. VSD generates good quality results by fine-tuning the 2D diffusion model. However, as we discussed in the main paper, it hurts the 2D diffusion model's comprehension capability to numerous text prompts, leading to mode collapse when the size of text prompts is extended. Without changing the diffusion prior, our proposed ASD can achieve the same high quality results as  VSD.

We also ablate the setting of $\Delta t$ in this experiment. We see that if we set $\Delta t = 0$, it leads to a noisy pattern similar to CSD. By setting it as a fixed interval, \eg, $\Delta t = \eta T_{\mathrm{max}}$, it would result in poor texture or geometry, such as the panda in Fig.~\ref{fig:2D_toy_exps_supp}. By setting $\Delta t$ relevant to $t$ as $\Delta t = \eta(t - T_{\mathrm{min}})$, the results can be much improved. Finally, the results are further enhanced by randomly sampling $\Delta t$ via $\Delta t \sim \mathcal{U}\left[0, \eta\left(t-T_{\min }\right)\right]$. The detailed explanations can be found in Sec.~\ref{sec:noise_prediction_behavior} of the main paper.

\begin{figure}[!t]
    \centering
    \includegraphics[width=\textwidth]{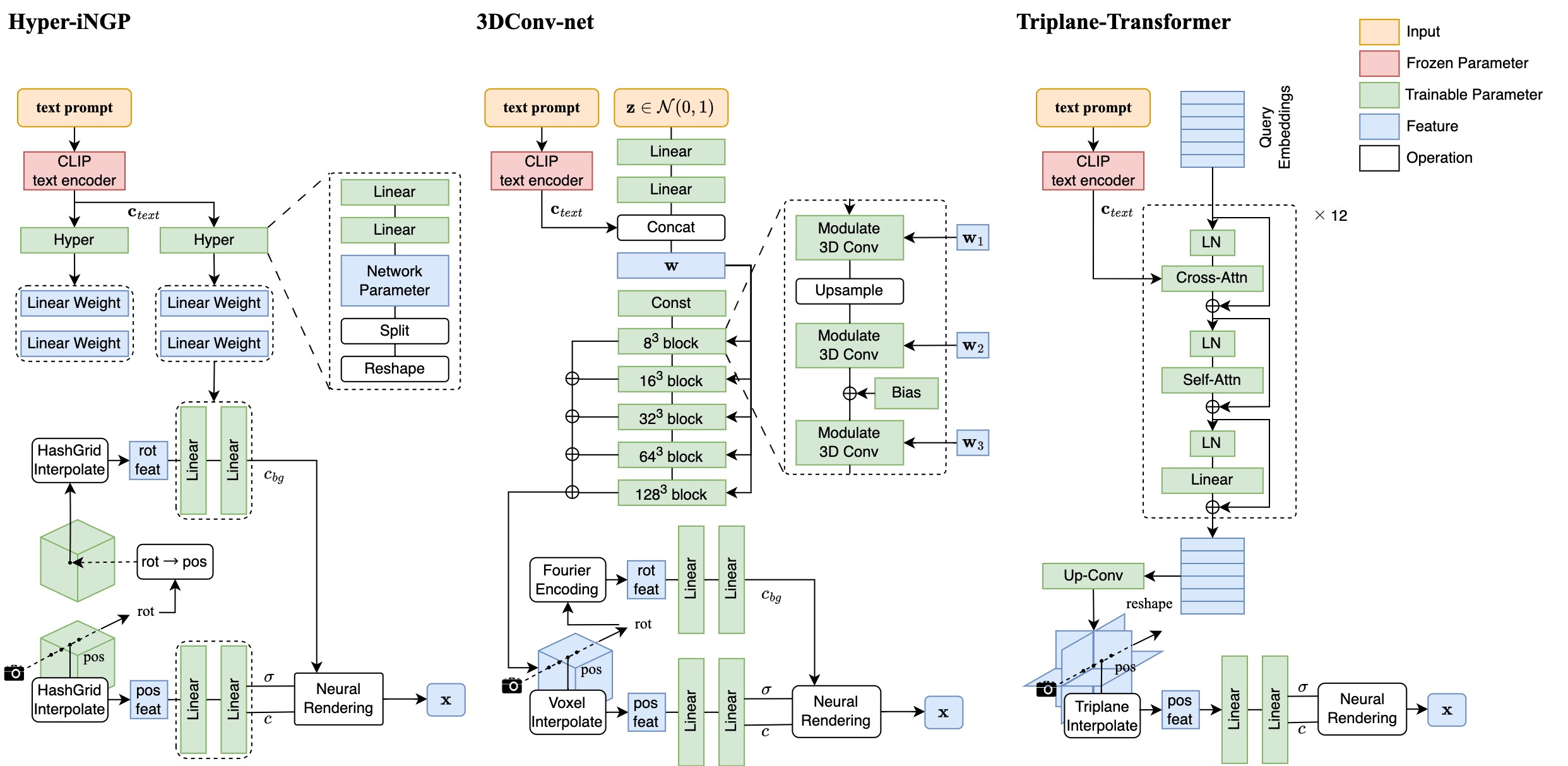}
    \caption{The network architecture and rendering scheme of  Hyper-iNGP(left), 3DConv-net(middle) and Triplane-Transformer (right)}
    \label{fig:network_supp}
\end{figure}

\section{More 3D Generator Architecture Details}
\label{sec:more_3D_generator_architecture_details}

\textbf{Hyper-iNGP}. We replicate the hypernetwork design from ATT3D~\cite{lorraine2023att3d}, integrating it with iNGP~\cite{muller2022instant} to achieve prompt-amortized text-to-3D synthesis. As illustrated in Fig.~\ref{fig:network_supp}, the hypernetwork projects text prompt embeddings into the weights of linear layers. The HashGrid representation~\cite{muller2022instant} encodes sample points independently, which are then transformed by the hypernetwork-parameterized linear layers into prompt-specific color $c$ and density $\sigma$. Following ATT3D~\cite{lorraine2023att3d}, another hypernetwork is implemented to create a prompt-specific background. The ray direction is encoded into a separate HashGrid, which is then projected to the background color $c_{bg}$, facilitating the creation of high-resolution backgrounds. The spectral normalization~\cite{miyato2018spectral} can be optionally turned on to stabilize the training with SDS~\cite{poole2022dreamfusion}.

\textbf{3DConv-net}. As illustrated in Fig.~\ref{fig:network_supp}, our 3DConv-net mirrors the StyleGAN2 model~\cite{karras2020analyzing}, using modulated convolutions to upscale features directed by the latent code $\mathbf{w}$, which is conditioned on Gaussian noise $\mathbf{z} \sim \mathcal{N}(0,1)$ and the text prompt embedding as in text-driven 2D GANs~\cite{sauer2023stylegan}. Transitioning from 2D to 3D, we substitute StyleGAN2's components with their 3D alternatives, modulated by $\mathbf{w}$. The network up-samples a $4^3$ dimensional voxel to $128^3$ dimension. For quicker convergence, we add 3D bias within blocks for processing voxels with the dimension from $8^3$ to $64^3$. Rendering is accomplished by interpolating voxel features to determine the color and density of each point along the rays. A background module is incorporated as well.

\textbf{Triple-Transformer}. Recently, the Transformer~\cite{vaswani2017attention} architecture has gained popularity in 3D generation tasks for its scalability, especially in data-driven methods~\cite{hong2023lrm,xu2024instantmesh,wei2024meshlrm,zou2023triplane,xu2024grm,xu2023dmv3d,tochilkin2024triposr,liu2023unidream,li2024m}. However, it has not been applied in recent score-distillation-based methods yet~\cite{li2023instant3d,qian2024atom,xie2024latte3d}. In this paper, we conduct experiments to explore the performance of Transformer architecture in score-distillation-based text-to-3D generation. As shown in Fig.~\ref{fig:network_supp}, we employ 12 Transformer layers, each comprising self-attention, cross-attention, and feed-forward networks. The text prompt is first processed by the CLIP text encoder and then fed into the cross-attention to set the condition. The query embeddings are passed through these layers, and then reshaped and up-sampled to form a triplane, which is an efficient 3D representation~\cite{chan2022efficient}.

\textbf{Rendering}. For prompt-specific optimization,  we use the volume rendering in NeRF~\cite{wang2023prolificdreamer} and keep the configuration in prior arts~\cite{wang2023prolificdreamer}. For prompt-amortized training, we implement VolSDF~\cite{yariv2021volume}, which uses 64 sample points for coarse sampling and 256 sample points for fine sampling~\cite{mildenhall2021nerf}. We found that keeping the  mean absolute deviation fixed to be 30 can achieve good results. We render $64 \times 64$ resolution for 3DConv-net and $256 \times 256$ for Hyper-iNGP in the whole training period. 

\section{More  Details about Corpus}
\label{sec:more_corpus_details}
In this work, we utilize five corpora to assess our ASD for prompt-based text-to-3D generation. Apart from MG15~\cite{lin2023magic3d}, DF415~\cite{poole2022dreamfusion}, AT2520~\cite{lorraine2023att3d} and DL17k~\cite{li2023instant3d}, we also provide the CP100k corpus. CP100k consists of 100k corpus for training and 1k corpus for test, which are sampled from Cap3D~\cite{luo2024scalable}.


\section{More Implementation Details}
\label{sec:more_implementation_details}


\textbf{Prompt-specific Text-to-3D}. Our code is based on the open-source Text-to-3D codebase~\cite{threestudio}. We follow the configuration in ProlificDreamer~\cite{prolificdreamer_2d} in specifying the parameters, including the training iterations, optimizer, batch-size and learning rate. All experiments are conducted on one Nvidia V100 GPU.

\textbf{Prompt-amortized Text-to-3D}. The experiments for prompt-amortized text-to-3D are conducted on 8 Nvidia A6000 GPUs, with a per-GPU batch size of 1. Training on MG15, DF417, AT2520, DL17k and CP100k requires 50k, 100k, 50k, 200k and 300k iterations, respectively.

\textbf{2D Diffusion Guidance}. For 2D experiments, utilizing the diffusion model~\cite{sdv2.1base} with $T=1000$ timesteps, we adhere to the existing protocol~\cite{prolificdreamer_2d} by setting $T_{\mathrm{min}}=20$ and $T_{\mathrm{max}}=980$. 
In the 3D experiments, we adopt the approaches in~\cite{wang2023prolificdreamer} and ~\cite{shi2023mvdream}, where $T_{\mathrm{max}}$ is progressively reduced from $980$ to $500$ to enhance the quality of generation outputs. We start with a higher $T_{\mathrm{min}}$ and decrease it linearly from $500$ to $20$, which helps to mitigate the Janus issue, as adopted in~\cite{armandpour2023re}. Additionally, when Stable Diffusion is used as the 2D diffusion model, we employ the Perp-neg strategy~\cite{armandpour2023re} to further address the Janus problem.




\end{document}